\documentclass[final,5p,times,twocolumn]{elsarticle}
\usepackage{amsmath,amssymb,graphicx,array,textcomp,adjustbox,placeins,float}
\usepackage[table]{xcolor}
\usepackage[hidelinks]{hyperref}
\hypersetup{
  pdftitle={Reconstructed holograms and explanation-aware evaluation for low-cost computational pollen analysis in veterinary cytology},
  pdfauthor={}, pdfsubject={}, pdfkeywords={}, pdfcreator={}, pdfproducer={}
}
\makeatletter\let\Hy@UseMaketitleInfos\relax\makeatother
\usepackage{lineno}
\newcolumntype{C}[1]{>{\centering\arraybackslash}m{#1}}
\newcommand{\tableheader}[2]{\begin{tabular}[c]{@{}c@{}}\textbf{#1}\\\textbf{#2}\end{tabular}}
\newcommand{\tablecell}[1]{\shortstack[c]{#1}}
\newcolumntype{+}{!{\vrule width 2pt}}
\newlength\savedwidth

\journal{Results in Engineering}
\begin{document}
\begin{frontmatter}
\title{Reconstructed holograms and explanation-aware evaluation for low-cost computational pollen analysis in veterinary cytology}
\author[edi]{Swarn Warshaneyan\corref{cor1}}
\ead{swarn.warshaneyan@gmail.com}
\author[edi]{Joial Danyal}
\author[lu]{Bla\v{z} Cugmas}
\author[lu]{Mindaugas Tamo\v{s}i\={u}nas}
\author[lu]{Edgars Kviesis-Kipge}
\author[edi]{Kirishanth Manivannan}
\author[edi]{Roberts Kadi\c{k}is}
\affiliation[edi]{organization={Institute of Electronics and Computer Science (EDI)}, city={Riga}, country={Latvia}}
\affiliation[lu]{organization={Faculty of Science and Technology, University of Latvia}, city={Riga}, country={Latvia}}
\cortext[cor1]{Corresponding author.}
\begin{abstract}
Automated pollen analysis supports veterinary cytology workflows, but conventional brightfield microscopy is costlier and mechanically more complex than lens-less digital in-line holographic microscopy. We evaluate whether reconstructed holograms can narrow this performance gap and whether model explanations remain reliable under the modality change. A silicone-fixed dataset of six pollen species was imaged by brightfield and holographic microscopy. Raw holograms and two published reconstruction types, p-type single back-propagation and r-type iterative phase retrieval, were evaluated with YOLOv26s detection and MobileNetV4 classification after anchor-based annotation transfer. Six attribution methods were assessed for spatial grounding and faithfulness using the Attribution Health Inspection and Repair (AHIR) protocol, which first tests model brittleness under weak noise and then corrects attribution-map granularity when needed. Brightfield achieved 0.6890 mAP50-95 (0.8865 mAP50) for detection and 0.9687 macro-F1 (0.9705 accuracy) for classification. Reconstructed holograms substantially narrowed the gap with a task-dependent split: p-type was strongest for detection at 0.5324 mAP50-95 (0.8229 mAP50), while r-type was strongest for classification at 0.7695 macro-F1 (0.7866 accuracy), both far above raw-hologram baselines. Activation-based explanations were strongly localized on true grains wherever evaluated, and region-based methods retained roughly 60 to 80\% of their faithfulness under holography. The evaluated holographic detector was highly brittle to weak perturbations, causing deletion-based evaluation to saturate while insertion remained informative. Pixel-level gradient explanations also collapsed toward the random floor, yet spatial smoothing restored p-type gradient faithfulness from 0.05 to 0.51 across all evaluated detections in the dedicated XAI subset. On holographic classification crops, perturbation-based explanations remained faithful while gradient-based methods fell below their random floor. Reconstruction therefore improves low-cost holographic pollen analysis, while explanation-aware evaluation distinguishes genuine attribution failure from artifacts caused by model brittleness and map granularity.
\end{abstract}
\end{frontmatter}
\section{Introduction}
Pollen grains are among the most clinically relevant aeroallergens in both human and veterinary medicine. In dogs, canine atopic dermatitis is a common, genetically predisposed, pruritic skin disease whose long-term management depends on identifying the environmental allergens that trigger it~\cite{hensel2015cad}. That identification is difficult and resource-intensive: intradermal testing is regarded as the diagnostic reference standard but must be performed by a veterinary dermatologist, and intradermal and serological tests show limited concordance, so cytological examination and careful case work-up remain central to routine practice~\cite{hensel2015cad, vetallergy2025}. Reliable recognition of pollen in cytological and environmental samples therefore supports allergen monitoring and diagnosis.

Traditionally, pollen is identified by expert palynologists under a light microscope, a process that is slow, labour-intensive, dependent on scarce expertise, and often resolved only to genus or family level~\cite{palyno2023}. Automation with deep learning has made optical pollen classification highly accurate, with convolutional neural networks reaching high accuracy on curated optical datasets~\cite{zolfaghari2024, rostami2023}. This success, however, rests on conventional optical/brightfield microscopy, which relies on comparatively expensive and mechanically complex instrumentation.

Digital in-line holographic microscopy (DIHM) is an attractive lower-cost alternative. In a lensless configuration it provides label-free imaging over a large field of view with a compact, inexpensive optical layout~\cite{kim2010dhm}. These properties make DIHM the desired modality for accessible, near-point-of-care pollen analysis, with optical/brightfield microscopy serving only as the performance baseline against which DIHM is judged. The difficulty is that in-line holograms are degraded by the twin-image artifact and lack the direct visual contrast of brightfield images, which makes them harder for automated models. Reconstruction and phase-retrieval methods, increasingly aided by deep learning, are used to recover clearer images from raw holograms~\cite{rivenson2018}. In our earlier studies, deep-learning detection and classification on raw holograms performed far below the optical baseline, even after dataset expansion and generative augmentation~\cite{warshaneyan2025ieee, warshaneyan2026sivp}.

This paper addresses that gap with a derived four-modality analysis dataset and computational pipeline. The samples are prepared with silicone fixation, previously shown to yield more analysis-friendly whole-slide images than adhesive tape~\cite{cugmas2024fixation}, and each raw hologram is accompanied by two reconstructions (a p-type single-backpropagation image and an r-type iterative Gerchberg-Saxton image~\cite{cugmas2026visual}), giving four effective imaging modalities from two strict acquisition modalities. We transfer optical annotations to the holograms through an improved anchor-based registration, evaluate object detection with YOLOv26s and object classification with MobileNetV4, and quantify how closely each holographic modality approaches the optical baseline.

Since strong scores by themselves do not prove that the holographic models are focusing on the pollen grains rather than on reconstruction artifacts or on the background structure, we also subject the trained models to a quantitative explainability analysis. Our contributions are five in number: (i) a derived four-modality annotated dataset spanning optical, raw holographic, p-type and r-type images; (ii) an anchor-based cross-modality registration which relies on the reconstruction most similar to brightfield as a bridging medium in order to transfer the optical annotations to all the holographic counterparts; (iii) a matched two-model detection-and-classification pipeline which greatly reduces, although not completely eliminates, the model performance gap between optical and holographic sides; and (iv) a quantitative explainable-AI evaluation that checks whether saliency-based explanations remain localized, faithful, and consistent across both the optical and the reconstructed holographic modalities, resulting in two transferable insights: a prominent perturbation brittleness of the holographic recognition models and a granularity mechanism which explains and remedies the apparent failure of gradient-based explanations; and (v) a reusable two-step Attribution Health Inspection and Repair (AHIR) protocol consisting of a noise gate that determines when faithfulness scores can be interpreted and a granularity correction that fixes pixel-level attributions, with the protocol having been validated using published detector-specific metrics.

\section{Related work}
\subsection{Automated pollen recognition}
Early automated pollen recognition relied on hand-crafted brightness- and shape-based descriptors combined with classical machine learning~\cite{rodriguez2004, goncalves2016}. The field has since shifted to deep learning: convolutional neural networks and transfer learning from pretrained backbones now dominate optical pollen classification~\cite{rostami2023, zolfaghari2024}, with applications ranging from tree-species identification~\cite{minowa2022} to multifocus pollen detection~\cite{gallardo2024}. Within veterinary medicine specifically, our collaborators established automated classification of veterinary-relevant pollens~\cite{cugmas2024nap} and showed that silicone fixation yields whole-slide images better suited to automated analysis than adhesive tape~\cite{cugmas2024fixation}. These works operate almost exclusively on optical/brightfield images, which motivates the present focus on holographic modalities.

\subsection{Digital holographic and lensless microscopy}
DIHM reconstructs quantitative, label-free images from recorded interference patterns and has a well-established theoretical basis~\cite{kim2010dhm}. Lensless in-line implementations are especially suited to low-cost, portable imaging, and deep learning has been applied to lensless microscopic imaging and reconstruction~\cite{grantjacob2022}. Our collaborators have investigated lensless on-chip microscopy for high-fidelity hologram reconstruction of pollen samples~\cite{murovec2024nap} and demonstrated visual classification of allergenic pollen from iteratively reconstructed lensless DIHM images~\cite{cugmas2026visual}, the pipeline from which the present dataset originates.

\subsection{Hologram reconstruction and twin-image suppression}
The central obstacle in in-line holography is the twin-image artifact, which arises from the loss of phase information and degrades both amplitude and phase reconstructions. Classical remedies include iterative phase-retrieval algorithms such as Gerchberg-Saxton~\cite{gerchberg1972}, which underlies the r-type reconstructions used here, whereas a single angular-spectrum backpropagation yields the simpler p-type images~\cite{cugmas2026visual}. More recent approaches use deep learning to recover clearer images, either by learning the phase-retrieval mapping directly~\cite{rivenson2018} or by suppressing twin-image noise with dedicated networks~\cite{utirnet}, and by inter-modality learning that enhances lensless holographic images using conventional microscope images as reference~\cite{intermodality2020}.

\subsection{Cross-modality translation and generative models}
Rather than transferring optical annotations to the holograms by geometric registration, the route adopted here, an alternative is to translate images directly between modalities. Unpaired image-to-image translation with cycle-consistent adversarial networks~\cite{cyclegan2017} has been used for cross-domain synthesis, and Wasserstein GANs with spectral normalization~\cite{wgan2017, spectralnorm2018} served as generative augmentation in our prior work~\cite{warshaneyan2026sivp}. Generative models have also been studied under high-dimensional, low-sample-size biological conditions, where data scarcity can materially affect model stability and generative fidelity~\cite{ghosh2026hdlss}. Related veterinary-imaging research on cross-modality virtual staining evaluated CycleGAN and ultimately discarded it in favour of a paired-supervision approach~\cite{viskere2026vs}. Consistent with these findings, our own attempt to translate holograms directly into the optical domain was defeated by the large resolution disparity between the modalities, which motivated the registration-based approach adopted here.

\subsection{Object detection and classification architectures}
We use YOLOv26s, the small variant of the end-to-end, non-maximum-suppression-free YOLO26 generation~\cite{yolo26, jocher2026yolo26}, and MobileNetV4~\cite{qin2024mobilenetv4}, comparing its convolutional and hybrid variants. Our earlier studies used YOLOv8 and MobileNetV3L on the previous dataset~\cite{warshaneyan2025ieee, warshaneyan2026sivp}.

\subsection{Explainable AI for detection and classification}
Post-hoc explanation methods for convolutional models fall into three broad families. Class-activation approaches derive saliency maps from internal activations, beginning with class activation mapping (CAM)~\cite{zhou2016cam} and Grad-CAM~\cite{selvaraju2017gradcam} and refined by Layer-CAM~\cite{jiang2021layercam}; extensions adapt them to object detectors, including Gaussian-localized variants~\cite{nguyen2023gcame} and detection-score targets~\cite{kirchknopf2022yolo}. Perturbation approaches probe the model by masking the input: systematic occlusion~\cite{zeiler2014occlusion}, randomized masks in Randomized Input Sampling for Explanation (RISE)~\cite{petsiuk2018rise} and its detector adaptation D-RISE~\cite{petsiuk2021drise}, and segment-based masking in the Morphological Fragmental Perturbation Pyramid (MFPP)~\cite{yang2020mfpp}. Gradient approaches attribute the prediction to input pixels, from raw and input-weighted gradients~\cite{shrikumar2016ixg} to Integrated Gradients~\cite{sundararajan2017ig} and noise-averaged variants such as SmoothGrad~\cite{smilkov2017smoothgrad}. Because saliency maps can be visually persuasive yet unfaithful, quantitative evaluation has become standard: deletion and insertion curves measure whether map-ranked pixels actually drive the prediction~\cite{petsiuk2018rise, zhang2021groupcam}, energy-based pointing measures whether map mass falls on the object~\cite{wang2020scorecam}, and model-randomization sanity checks test whether maps depend on the learned weights at all~\cite{adebayo2018sanity}. Explainability is increasingly expected in biomedical imaging applications~\cite{vandervelden2022xai}, yet neither of our prior pollen studies included any explanation analysis, and to our knowledge no study has examined whether saliency-based explanations survive the transition from optical microscopy to holographic imagery, and the present work addresses both points.

Three strands of recent work frame our contribution. For object detectors, black-box explainers that build on D-RISE with hierarchical or multi-level segmentation masks have been proposed, namely BODEM~\cite{moradi2024bodem}, D-CLOSE~\cite{truong2024dclose} and D-MFPP~\cite{andres2024dmfpp}. The latter study also introduced D-Deletion, a deletion metric that combines faithfulness with localization for detectors, which we adopt as the closest published comparator to our faithfulness axis. In biomedical imaging, where saliency maps are still used mostly for qualitative inspection~\cite{fayyaz2025gradcam}, quantitative evaluation of interpretability methods has begun to appear: attribution faithfulness has been evaluated across medical image datasets~\cite{lamprou2024faith}, attribution maps have been used as the instrument for comparing networks trained on label-free microscopy~\cite{scodellaro2025xai}, and cross-microscope domain shift has been studied for blood-smear analysis~\cite{ilyas2024malaria}. For pollen specifically, explainable AI has been applied to classification from single-particle scattering and fluorescence signals rather than images~\cite{brdar2023pollenxai}, and image classification has been advanced with architecture design~\cite{zu2024swint}. To our knowledge no prior study holds the specimen fixed, varies the imaging modality, and evaluates whether explanations of both a detector and a classifier survive the change.

\section{Data}

\subsection{Image modalities and sample preparation}
\begin{figure*}[!t]
\centering
\includegraphics[width=0.9\textwidth]{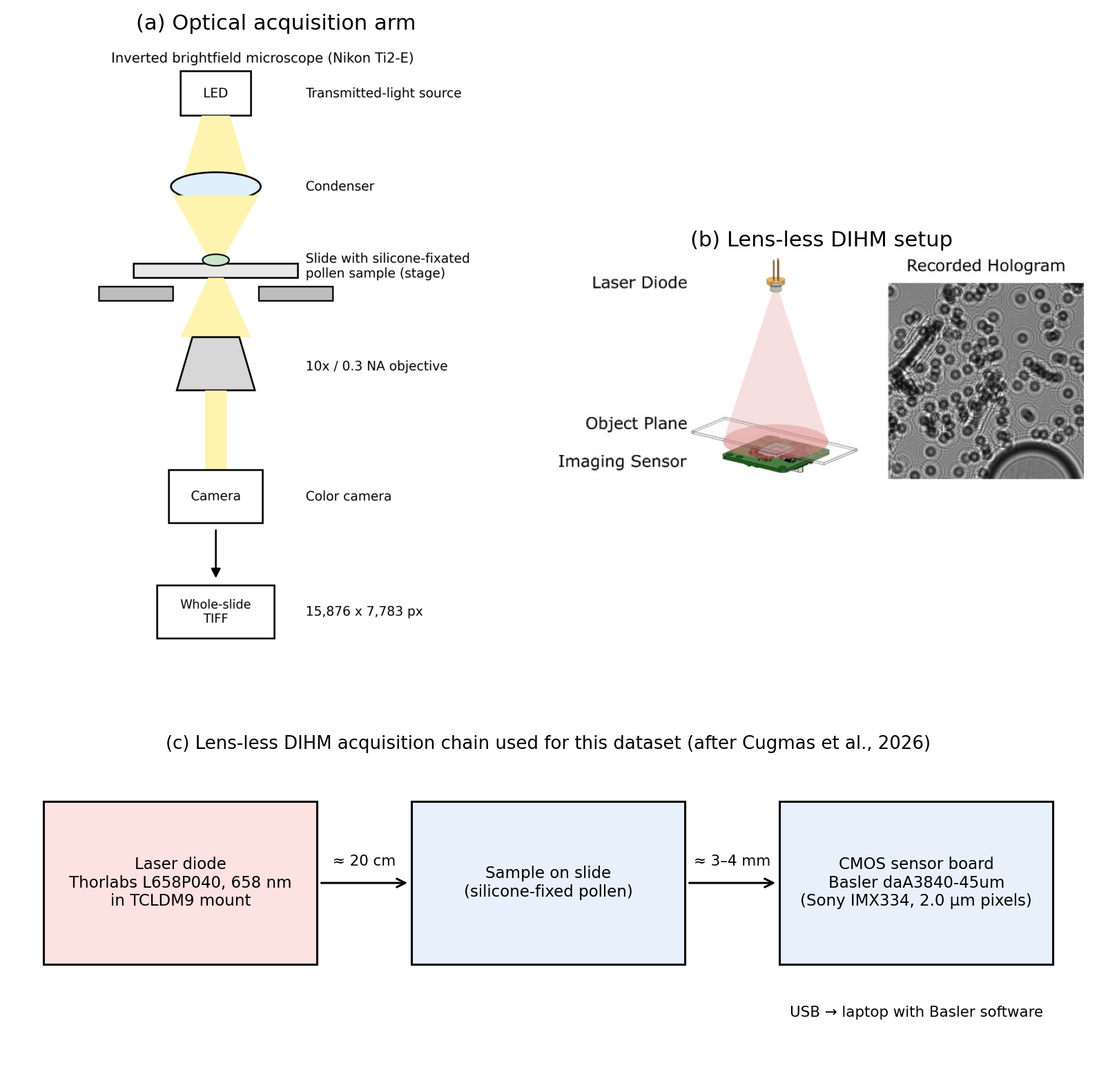}
\caption{\textbf{Imaging setups for the two strict acquisition modalities.}
(a)~Optical acquisition arm of the inverted brightfield microscope (Nikon Ti2-E).
(b)~Lens-less DIHM setup schematic with an example hologram, reproduced from~\cite{cugmas2026visual}
under the CC BY 4.0 license. (c)~Lens-less acquisition chain used for this dataset, after~\cite{cugmas2026visual}.}
\label{fig:setup}
\end{figure*}

The central premise of this work is that digital in-line holographic microscopy (DIHM) is the desired imaging modality, because it is substantially cheaper and mechanically simpler than conventional optical/brightfield microscopy, whereas optical/brightfield microscopy serves only as the reference baseline against which DIHM performance is judged. All experiments are therefore framed as attempts to bring DIHM-based detection and classification performance as close as possible to the optical/brightfield baseline. The two acquisition setups are illustrated in Fig~\ref{fig:setup}.

\begin{figure*}[!t]
\centering
\includegraphics[width=0.9\textwidth]{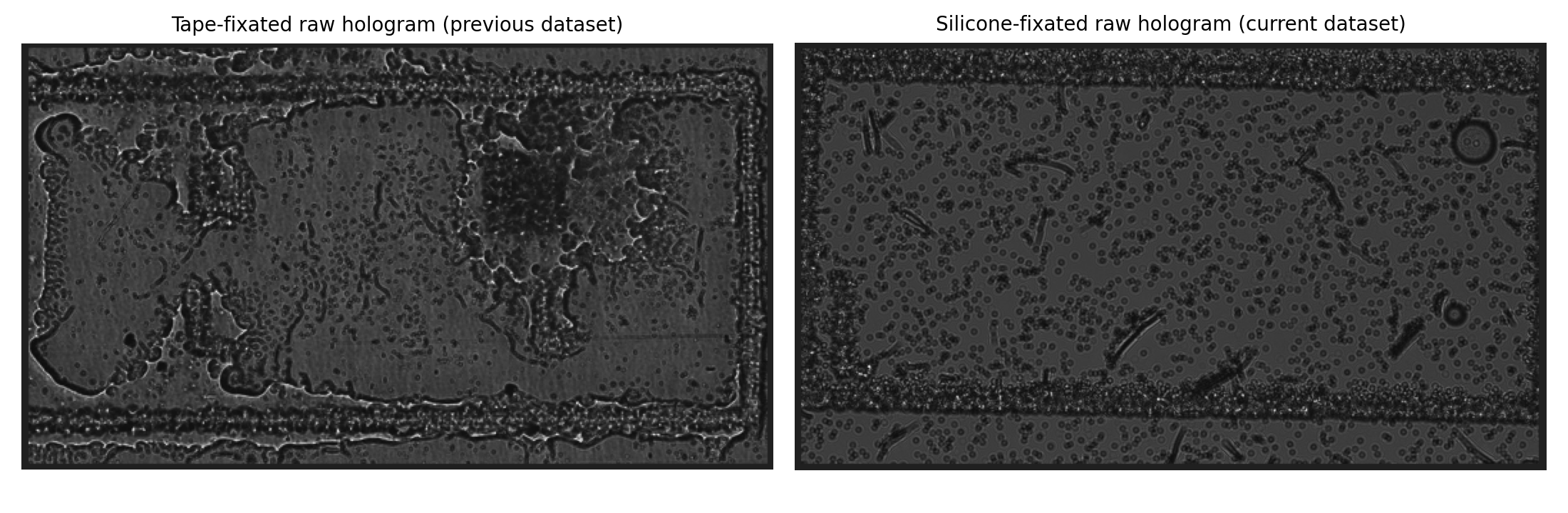}
\caption{\textbf{Fixation methods compared on raw holograms.}
A tape-fixated raw hologram from the previous dataset~\cite{warshaneyan2026sivp} and a
silicone-fixated raw hologram from the current dataset. The silicone preparation yields a
visually cleaner field.}
\label{fig:fixation}
\end{figure*}

The dataset was acquired and shared by the University of Latvia group, co-authors of this work, and originates from the sample and imaging pipeline described in~\cite{cugmas2026visual}. Pure, dry pollen of six clinically relevant species was embedded in a silicone mounting medium (PlatSil SiliGlass) and sealed under coverslips, replacing the adhesive-tape fixation used in our earlier studies~\cite{warshaneyan2025ieee, warshaneyan2026sivp}. Silicone fixation had previously been shown to yield higher automated classification accuracy than tape~\cite{cugmas2024fixation}. Fig~\ref{fig:fixation} illustrates the visual difference between the two fixation approaches on raw holograms.

 Each species was presented on ten slides, giving sixty slides in total. The six species are those of the source study~\cite{cugmas2026visual} (timothy grass, common ragweed, silver birch, common alder, olive tree, and hazel). They are referred to throughout by the dataset class codes A1 to A5 and A7, because the code-to-species mapping was neither available to us nor required by any step of the pipeline.

Two strict acquisition modalities were used: optical/brightfield and holographic. Optical images were captured on a Nikon Ti2-E inverted microscope (10x objective with 0.3 numerical aperture (NA), color camera) as 15{,}876~$\times$~7{,}783 pixel TIFFs
, while holograms were captured on a custom lens-less DIHM with a red laser diode (658~nm, L658P040, Thorlabs)~\cite{cugmas2026visual} as 3{,}840~$\times$~2{,}160 pixel TIFFs~\cite{cugmas2026visual}. From each raw hologram, two computational reconstructions were additionally produced: a p-type image obtained by a single angular-spectrum backpropagation to a manually selected focus distance, and an r-type image obtained by iterative Gerchberg-Saxton phase retrieval at that distance~\cite{cugmas2026visual}. This yields four effective modalities used in the experiments (optical/brightfield, raw hologram, p-type hologram, and r-type hologram), although only two strict acquisition modalities exist.

\begin{figure*}[!t]
\centering
\includegraphics[width=0.9\textwidth]{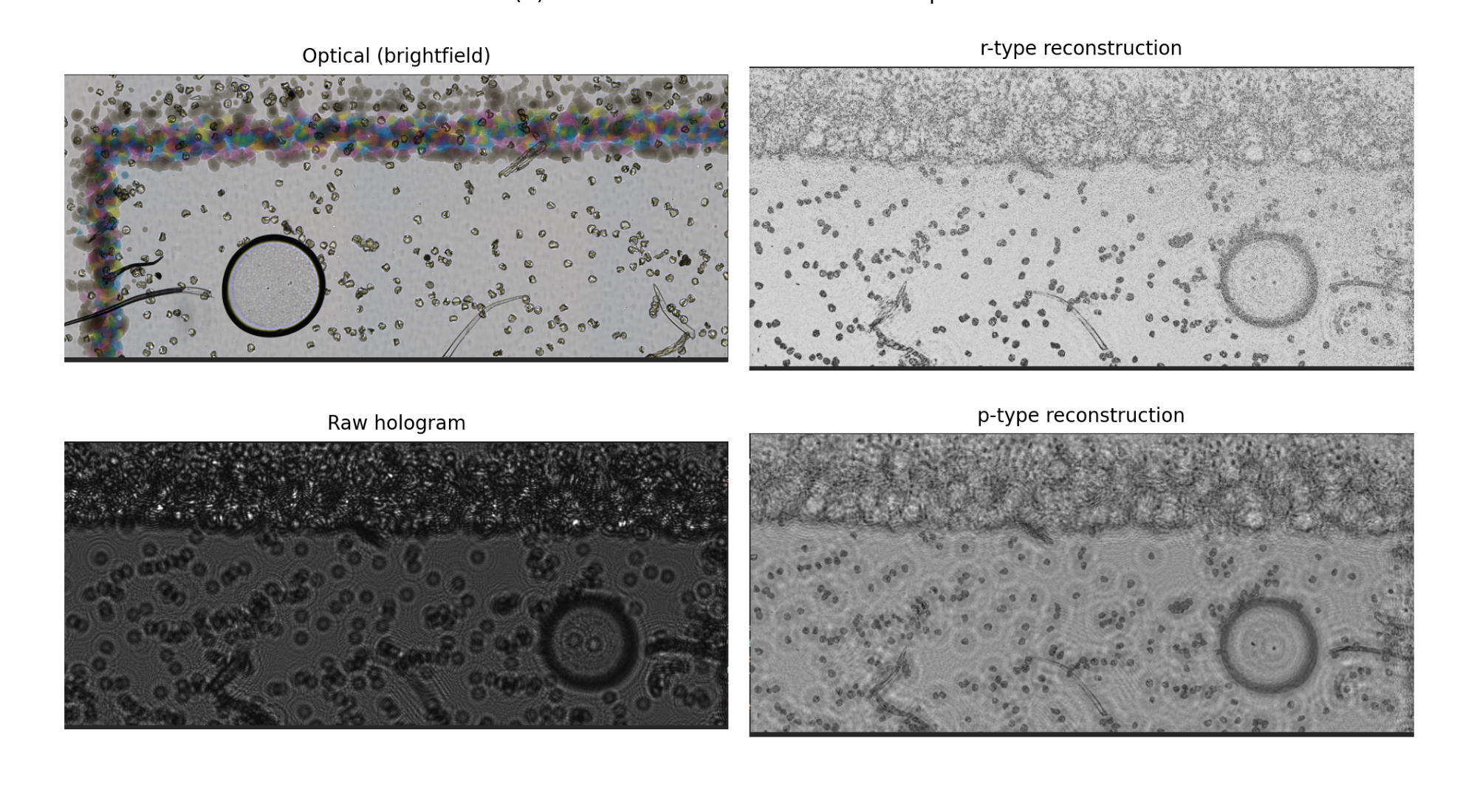}
\caption{\textbf{The four effective imaging modalities at grain scale.} Zoomed views of the same field for optical/brightfield, raw hologram, p-type (single-backpropagation) and r-type (Gerchberg-Saxton iterative) reconstructions; whole-field views are given in Fig~S5.}
\label{fig:modalities}
\end{figure*}

Representative images of all four modalities for one sample are shown in Fig~\ref{fig:modalities}.

\subsection{Annotation and labeling}
Object annotations were produced on the full-resolution optical/brightfield images and stored in the LabelMe JSON format. The holographic images cannot consume these optical annotations directly and instead receive them through the alignment step described below. Detection labels treat every pollen grain as a single merged object class, so the per-class labels A1 to A5 and A7 denote only the source species of each grain rather than a detection target.

Training and validation labels were generated automatically by the class-agnostic YOLOv8s detector of our previous study~\cite{warshaneyan2026sivp}, which was trained on the earlier tape-fixated dataset, applied through sliced inference with greedy non-maximum merging and an intersection-over-smaller match metric (standard non-maximum suppression with intersection over union, IoU, was evaluated but gave less satisfactory results). The automatically detected objects were stored as polygons and converted to axis-aligned rectangles. The test subset was annotated manually and serves as the ground truth for every reported metric.

The sixty images were split at the slide level into eight training, one validation, and one test image per species (an 8:1:1 split), giving 48 training, 6 validation, and 6 test images.

\subsection{Image alignment and registration}
Because the optical and holographic modalities differ greatly in resolution (the optical images are roughly four times larger per side than the holographic images), transferring the optical annotations onto the holograms requires an accurate geometric registration between them, because residual misalignment would propagate into every downstream computation, as in other multimodal sensing pipelines~\cite{sayed2024deviations}. Our prior pipeline addressed this with a lightweight registration script whose residual inaccuracies were the original motivation for the bounding-box expansion step described later.

Motivated by related cross-modality image-translation work in veterinary imaging~\cite{viskere2026vs}, we first attempted to bypass registration by learning a direct holographic-to-optical translation with a cycle-consistent adversarial network~\cite{cyclegan2017}. The translations were unsatisfactory on this dataset, so we returned to the registration itself and improved on the lightweight script of the previous pipeline.

The improved registration exploits the fact that the three holographic variants (raw, p-type, r-type) are mutually co-registered, being different reconstructions of the same physical capture. The r-type reconstruction most closely resembles the optical/brightfield appearance, so it is used as an anchor: a partial-affine transform (rotation, uniform scale, translation) is estimated between the r-type hologram and the optical image, and the same transform is then reused for the raw and p-type holograms. Correspondences are established with scale-invariant feature-transform keypoints~\cite{lowe2004sift} computed on gradient-magnitude images, so that matching relies on shared edge structure rather than on raw intensity, which differs strongly between the modalities, and filtered with a ratio test and a random sample consensus (RANSAC)~\cite{fischler1981ransac} inlier criterion. All pairs registered successfully, and the holograms are warped into the optical reference frame so that the optical annotations apply directly.

\begin{figure*}[!t]
\centering
\includegraphics[width=0.9\textwidth]{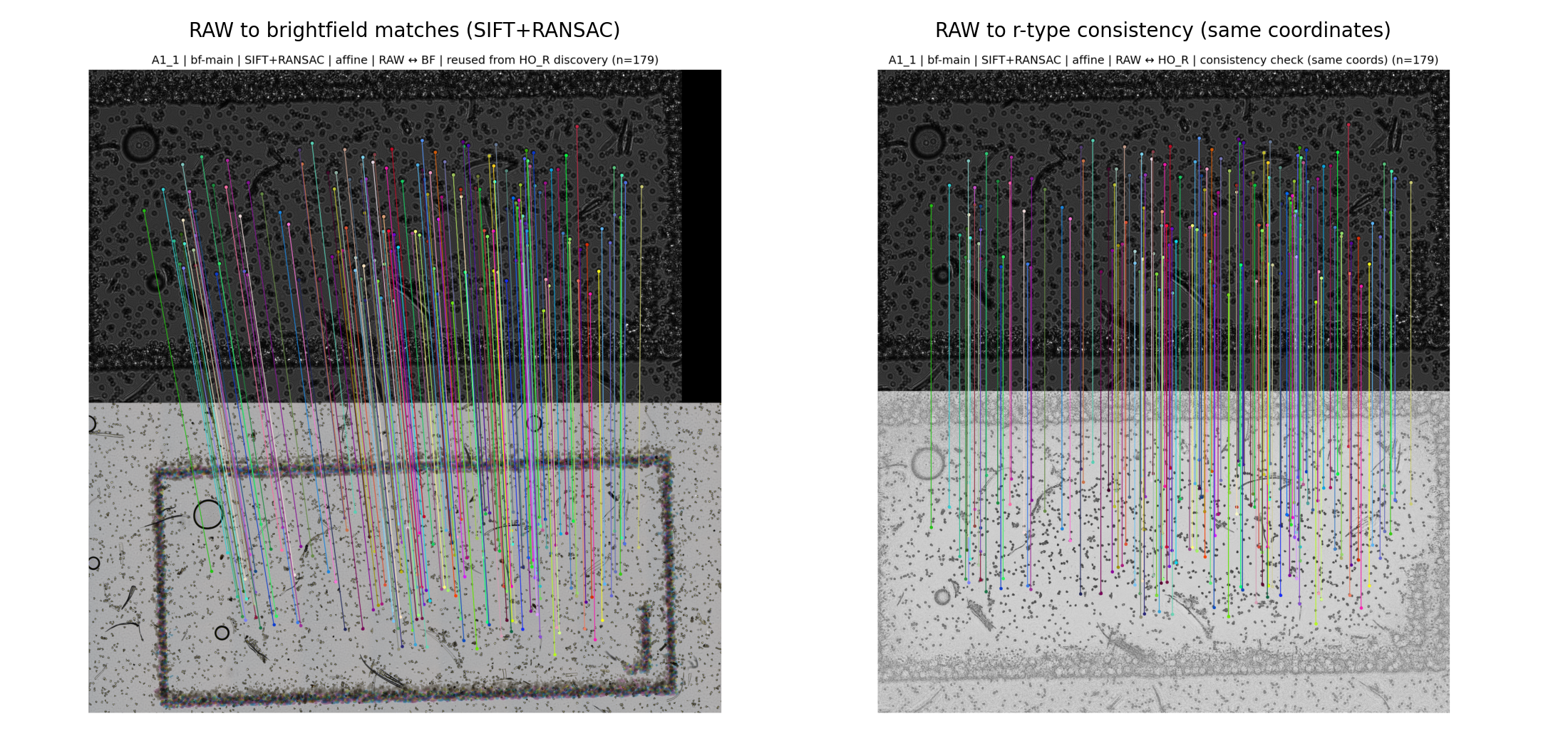}
\caption{\textbf{Anchor-based registration of the holographic modalities to the optical reference.}
Left: scale-invariant feature transform (SIFT) + RANSAC keypoint matches between the raw hologram and the brightfield reference under the
reused affine transform. Right: consistency check between the raw hologram and the r-type
reconstruction at identical coordinates. The black band on the right edge of the left panel is the
out-of-frame region produced by the size mismatch between the modalities.
The r-type reconstruction is registered to the optical image. The resulting partial-affine transform
is reused for the raw and p-type holograms.}
\label{fig:alignment}
\end{figure*}

A schematic of the anchor-based registration is shown in Fig~\ref{fig:alignment}.

\subsection{Blackening removal and object-instance accounting}
Warping the holograms into the optical frame leaves black borders wherever the smaller holographic field does not cover the optical field. These black regions were removed by cropping, which also discards the grains that fall inside them. The cropping step is therefore the point at which the holographic object counts diverge from the optical counts. Per-class retention after cropping ranged from 67.9\% to 73.1\% (Table~S1): overall, 227,128 optical instances are reduced to 161,755 holographic instances, a retention of 71.22\%.

A second, smaller divergence in object counts arises \emph{after} cropping, between the two downstream tasks. The detection pipeline tiles each cropped image into fixed-size 640-pixel patches, so a grain straddling a tile boundary can be split across tiles, whereas the classification pipeline extracts one crop per grain. The detection object count is therefore higher than the shared post-crop instance count (35,591 tiled detection objects against 29,591 post-crop instances for the test split), while the classification crop count matches the post-crop count exactly. The dataset splits at the image and instance level are listed in Table~S2.

\subsection{Bounding box area expansion}
The bounding-box expansion step, introduced in our prior work to compensate for registration inaccuracy, was retained here as an ablation. Crucially, it was applied only to the holographic classification crops (raw, p-type, and r-type), at 100\%, 125\%, and 150\% bounding box area, because the holographic classification results were the weakest part of the pipeline and thus the target for improvement. Optical/brightfield classification used 100\% bounding box area only, as it already performed close to the ceiling and did not require this remedy, and all object detection runs (every modality) likewise used 100\% only, and non-100\% detection was intentionally left out of scope to keep the reporting focused. Because the improved registration reduces the misalignment this step was designed to absorb, its benefit is no longer uniform, and the corresponding classification results (Table~S4) show a modality-dependent effect. The three expansion levels differ in retained instance count by at most a few tens of objects out of roughly 161{,}755 (under 0.1\%), so these differences are treated as negligible and are not tabulated separately. An illustration of the three expansion levels is given in Fig~S1.

\section{Experiments}

\subsection{Performance judgement criteria}
Following the framing above, every experiment is evaluated by how close the holographic result comes to the optical/brightfield baseline. Two metrics are reported per task, chosen for comparability with our prior papers. For object detection, mean average precision at an IoU threshold of 0.5 (mAP50) is reported for backward comparability and its stricter multi-threshold variant (mAP50-95) as the primary metric. For object classification, accuracy is reported for backward comparability and macro-F1 as the primary metric. All reported numbers are computed on the six-image manually verified test subset.

\subsection{Object detection training}
Object detection used YOLOv26s trained on the tiled images with all pollen grains merged into a single object class. Training ran for a maximum of 200 epochs with early stopping enabled; the detection runs reached the 200-epoch limit.

A single YOLOv26s model was additionally tested as a combined detect-and-classify baseline on raw holograms, trained for 400 epochs, and its degraded performance (reported in Results) motivated the two-model design used throughout the remaining experiments.

The full YOLOv26s training configuration is listed in Table~\ref{table:train_config_det}.
\begin{table*}[!t]
\centering
\caption{\textbf{YOLOv26s training configuration (identical recipe per modality).}}
\begin{adjustbox}{max width=\textwidth}\begin{tabular}{|C{2.1in}|C{2.1in}|}
\hline
\textbf{Setting} & \textbf{Value}\\ \hline
Epochs (early-stopping patience) & 200 (25)\\ \hline
Batch size / image size & 8 / 640\\ \hline
Optimizer & auto (SGD family), cosine LR, lr0 0.01, lrf 0.01\\ \hline
Momentum / weight decay & 0.937 / 0.0005\\ \hline
Warmup epochs & 3\\ \hline
Augmentation & mosaic 1.0 (off last 10 epochs), HSV 0.015/0.7/0.4, rotation 45$^{\circ}$, translate 0.1, scale 0.5, flips 0.5/0.5\\ \hline
Single-class mode / automatic mixed precision (AMP) / seed & on / on / 42\\ \hline
\end{tabular}\end{adjustbox}
\label{table:train_config_det}
\end{table*}

\begin{table*}[!t]
\centering
\caption{\textbf{MobileNetV4 Conv Medium two-phase training configuration (identical recipe per modality).}}
\begin{adjustbox}{max width=\textwidth}\begin{tabular}{|C{2.1in}|C{2.1in}|}
\hline
\textbf{Setting} & \textbf{Value}\\ \hline
Training image size & 256\\ \hline
Initialization / input normalization & ImageNet-pretrained backbone / ImageNet mean and standard deviation\\ \hline
Augmentation & horizontal and vertical flips and 90$^\circ$ rotations (each $p$~=~0.5); brightness/contrast jitter $\pm$0.10 ($p$~=~0.4)\\ \hline
Loss & class-weighted cross-entropy, label smoothing 0.1\\ \hline

Phase 1 optimizer & AdamW, LR 1$\times$10$^{-3}$\\ \hline
Phase 1 batch size (max epochs) & 512 (60)\\ \hline
Phase 2 learning rate & 5$\times$10$^{-5}$, reduce-on-plateau schedule (factor 0.5, patience 3, minimum 10$^{-7}$)\\ \hline
Phase 2 batch size (max epochs) & 104 (250)\\ \hline
Phase 2 mechanisms & batch-normalization (BN) statistics frozen; weighted sampler\\ \hline
Early-stopping patience & 15 (phase 1), 30 (phase 2)\\ \hline
Checkpoint selection & best validation epoch across phases (best epochs 31 and 45, optical run)\\ \hline
\end{tabular}\end{adjustbox}
\label{table:train_config_cls}
\end{table*}
\subsection{Object classification training}
Object classification used MobileNetV4 trained on the extracted per-grain crops, with a two-phase schedule (a frozen-backbone phase of up to 60 epochs with patience 15, followed by a fine-tuning phase of up to 250 epochs with patience 30). Phase 1 trains the classification head with the backbone frozen, and phase 2 fine-tunes the network with batch-normalization statistics frozen and a class-weighted sampler. The best-validation epochs of the two phases for the optical run were 31 and 45.

The full MobileNetV4 Conv Medium training configuration is listed in Table~\ref{table:train_config_cls}. The classification configuration is reported at a finer mechanism level than the detection configuration because the two pipelines differ in style: the YOLOv26s runs use the Ultralytics training pipeline, which automates most optimization behavior behind the settings listed in Table~\ref{table:train_config_det}, whereas the MobileNetV4 procedure is a custom two-phase pipeline in which these mechanisms are controlled explicitly.

The choice of MobileNetV4 variant was itself an experiment. Because YOLOv26s (a hybrid architecture) performed well for detection, a hybrid MobileNetV4 variant (which interleaves attention blocks with the convolutional backbone) was screened against the convolutional variant on the raw-hologram 100\% condition. The hybrid variant was less stable: unfreezing more than the final backbone block during the fine-tuning phase caused optimization collapse, and even its best stable configuration remained below the convolutional variant on both retained metrics, so MobileNetV4 Conv Medium was selected for all remaining classification experiments.

\subsection{Explainable AI evaluation}

Six attribution methods were evaluated, each producing a saliency map per decision. For detection, Layer-CAM-G computes a Layer-CAM map~\cite{jiang2021layercam} on an intermediate detector layer with a detection-score target~\cite{kirchknopf2022yolo} and a Gaussian localization term following G-CAME~\cite{nguyen2023gcame}. For classification, standard Layer-CAM is used. Occlusion slides a mean-color masking window and records the score drop~\cite{zeiler2014occlusion} (64/32-pixel window/stride at 640 pixels for detection; 32/16 at 256 for classification). D-RISE aggregates 250 randomized low-resolution masks weighted by the detection score~\cite{petsiuk2021drise}. MFPP-D applies multi-level superpixel masking~\cite{yang2020mfpp} (simple linear iterative clustering, SLIC, at 50, 150, and 300 segments, 84 masks per level) with D-RISE-style detection weighting. This construction is our independent implementation of the multi-level segmentation family also represented by D-CLOSE~\cite{truong2024dclose} and D-MFPP~\cite{andres2024dmfpp}, and we keep the MFPP-D name to make that lineage explicit. Input\,$\times$\,Gradient (IxG) multiplies each input pixel by its gradient~\cite{shrikumar2016ixg}, and Integrated Gradients (IG) averages gradients along a 50-step path from a baseline~\cite{sundararajan2017ig}. IG is evaluated for classification. All classification-side explanations are computed at the classifier training resolution of 256 pixels. A paired random-map baseline accompanies every method, and noise-averaged rescue variants (SmoothGrad and related estimators~\cite{smilkov2017smoothgrad}) were evaluated once under pre-declared protocols. Three methods are shared by both tasks and anchor the cross-task comparison: Layer-CAM (as Layer-CAM-G on the detector), Occlusion and IxG. D-RISE and MFPP-D are detector-specific black-box methods and run on detection only; Integrated Gradients runs on classification only. The detector-specific methods D-RISE and MFPP-D were scoped to the primary detection comparison between the optical baseline and p-type, the strongest holographic detection modality, so that on the detection side all five methods are evaluated on both optical and p-type images, while r-type detection was retained as a secondary cross-reconstruction check through Layer-CAM-G. Classification was evaluated on optical, r-type and p-type crops with all four classification methods, so that the classification-optimal r-type reconstruction can be compared both with the optical baseline and with p-type, which is optimal for detection. The resulting method-by-modality design therefore follows the task-dependent performance structure observed in the foundational experiments. The detection side receives the deeper treatment (five methods with both spatial and faithfulness evaluation), the classification side a compact one (four methods, primarily faithfulness). Detection also receives the deeper treatment because it sits upstream of classification: a grain missed or misexplained at the detection stage propagates through the entire pipeline, making detection the more consequential potential bottleneck.

Explanations are evaluated on the manually verified test subset along three axes. Faithfulness uses deletion and insertion curves~\cite{petsiuk2018rise}: pixels are removed or revealed in map-ranked order over 30 steps with mean-color fill, and the Over-all score (insertion minus deletion area under the curve, following Group-CAM~\cite{zhang2021groupcam}) summarizes both. The score target is the class probability for classification and the best-IoU-matched detection score for detection.

Spatial grounding uses the energy-based pointing game (EBPG)~\cite{wang2020scorecam} against the manual ground-truth boxes, reported together with a localization lift that divides the in-box energy fraction by the box's share of the image area, so that every value reads as a multiple of geometric chance; a seeded random-map null provides the empirical floor.

Sanity is assessed by model-randomization checks~\cite{adebayo2018sanity}, including a cascading variant that randomises progressively deeper detector stages. Detection-side evaluation covers every prediction matched to a ground-truth box at IoU~$\geq$~0.45 with confidence~$\geq$~0.20 (381 optical, 313 p-type, and 309 r-type detections for the spatial tables). Classification-side evaluation uses 42 crops per species and modality (252 optical, 248 r-type, and 248 p-type crops), scored against the predicted class with paired designs. The noise-gate validation uses its corresponding seeded per-arm cohorts, with cohort sizes reported in Table~\ref{table:gate}. All stochastic components use fixed seeds (42) and reproduce the reported values on repetition. The confidence threshold of 0.20 lies below the default prediction threshold so that marginal true detections enter the explained set. The matching threshold of IoU~$\geq$~0.45 tolerates grains truncated at 640-pixel tile boundaries, and matching explained detections to ground truth by IoU follows the D-RISE evaluation practice~\cite{petsiuk2021drise}. The classification subset was sampled at 42 crops per species to give each species equal weight in the explanation comparison and to prevent class frequency from dominating the aggregate results.

\subsection{Attribution Health Inspection and Repair (AHIR) protocol}
The AHIR protocol uses two steps, an inspection step (the noise gate) and a repair step (the granularity correction), to decide, per item, whether a faithfulness score is interpretable and, if it is, whether a failing pixel-level attribution can be repaired.
\begin{enumerate}
\item \textbf{Noise gate (inspection).} For each explained item, the clean model score $s_0$ (matched detection score, or class probability) is compared with the mean score $\bar{s}_\sigma$ under $N$ draws of Gaussian pixel noise of standard deviation $\sigma$, and the retention $\bar{s}_\sigma/s_0$ is recorded. Items with retention below 0.5 are flagged, and an arm in which every item collapses is declared not interpretable: its deletion-based faithfulness measures brittleness, not attribution quality. $\sigma$ is calibrated by a dose-response sweep (Results).
\item \textbf{Granularity correction (repair).} Pixel-level maps are smoothed with a Gaussian kernel ($\sigma$~=~4, 8 and 16 pixels) before scoring; the gain over the plain map is tested per item.
\end{enumerate}
Two controls accompany the protocol. Every attribution map is paired with a seeded random map on the same item, and faithfulness is reported as the excess of the map over its own random floor. For detection, the conclusions are cross-checked in the published D-Deletion metric~\cite{andres2024dmfpp} and its insertion-side counterpart, D-Insertion: D-Deletion is the area under the curve of $\max_j O_j\,\mathbb{1}[\mathrm{IoU}(d_t, d_j) > \gamma]$ as ranked pixels are removed, where $O_j$ is the objectness of proposal $j$ and $d_t$ the target box. We use $\gamma$~=~0.45 and 0.50. D-Insertion applies the same score as ranked pixels are revealed on a mean-colour canvas (higher is better), and D-Min-Insertion is the revealed fraction at which a proposal first exceeds $\gamma$ (lower is better). Bootstrap confidence intervals (4,000 resamples) and paired sign tests are used throughout.

Explanations were evaluated on a fixed, seeded subset of the manually verified test images: three 640-pixel patches per test image and modality (18 patches per modality), within which every matched detection was explained, together with the species-balanced classification subset described above. Sampling at the patch level preserves the local image context used by the detector and avoids selecting individual detections according to their prediction or explanation outcome. Complete inclusion of matched detections within each selected patch provides several hundred paired item-level observations per arm for the cross-modality explanation comparisons, which are evaluated with paired statistics and bootstrap confidence intervals.

We used Claude and ChatGPT for limited assistance in refining and debugging selected analysis scripts. All resulting code and outputs were reviewed and verified by the authors.

\section{Results}

All performance values reported in this section were obtained from the test subsets using the best checkpoint selected during training. Object detection and object classification were treated as separate tasks because they differ in model architecture, output structure and evaluation metrics. YOLOv26s was used for object detection, while MobileNetV4 Conv Medium, selected after comparison with MobileNetV4 Hybrid Medium, was used for crop-level object classification. For continuity with our previous optical-versus-holographic microscopy studies, mAP50 and overall accuracy are still reported alongside the current primary metrics, mAP50-95 for detection and macro-F1 for classification.

\begin{figure*}[!t]
\centering
\includegraphics[width=0.9\textwidth]{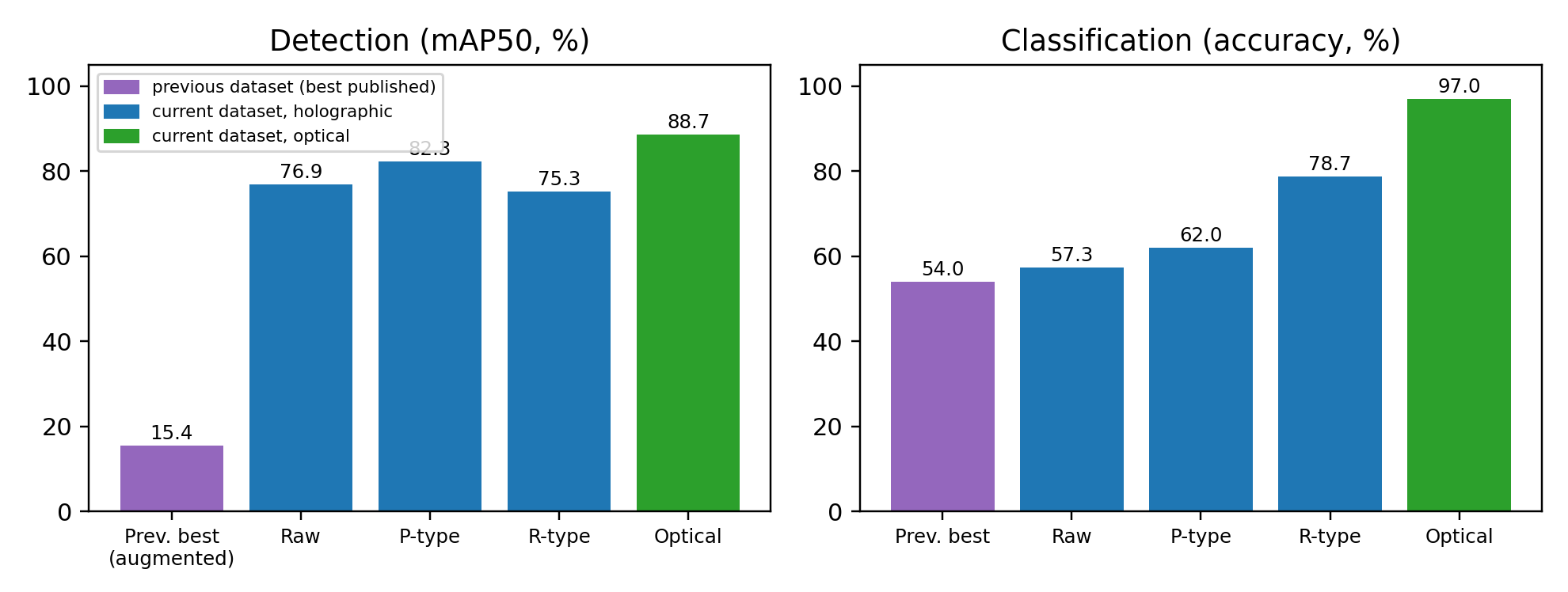}
\caption{\textbf{Narrowing of the optical-holographic gap relative to the previous dataset.}
Best previously published raw-hologram values (purple; detection includes synthetic-data
augmentation~\cite{warshaneyan2026sivp}) against the current results at 100\% bounding box
area. The two datasets differ in fixation, acquisition, and preparation, so the comparison
reflects the progress of the overall pipeline rather than a controlled benchmark.}
\label{fig:gap_progress}
\end{figure*}

The previous dataset contained optical (brightfield) and raw holographic images only, whereas the current dataset adds the p-type and r-type reconstructions described in the Data section. The primary objective of the present experiments was to determine whether any holographic modality could approach optical-image performance for either object detection or object classification. The strongest holographic detection results were obtained with p-type images, while the strongest holographic classification results were obtained with r-type images. Fig~\ref{fig:gap_progress} summarizes this progress.

A single YOLOv26s model trained to detect and classify at once (raw holograms, 100\% bounding box area) reached a matched-detection accuracy of 0.8334, but only 18,225 of 35,591 ground-truth objects were matched, and counting the misses as failures gives an all-object accuracy of 0.4268 and macro-F1 of 0.3324 (Table~S3), which motivated the two-model design (one model for detection and a separate one for classification).

The detection-only YOLOv26s results are summarized in Table~\ref{table:yolo_detection_results}.

\begin{table}[!t]
\centering
\caption{\textbf{YOLOv26s object detection performance on the test subset using best checkpoints.}}
\begin{adjustbox}{max width=\linewidth}\begin{tabular}{|C{1.25in}|C{0.70in}|C{0.65in}|C{0.65in}|}
\hline
\textbf{Modality} & \textbf{Box area} & \textbf{mAP50} & \textbf{mAP50-95}\\ \hline
Raw hologram & 100\% & 0.7694 & 0.4565\\ \hline
P-type hologram & 100\% & 0.8229 & 0.5324\\ \hline
R-type hologram & 100\% & 0.7528 & 0.4756\\ \hline
Optical microscopy & 100\% & 0.8865 & 0.6890\\ \hline
\end{tabular}\end{adjustbox}
\begin{flushleft}
Detection metrics are reported for the test subset using the best checkpoint.
\end{flushleft}
\label{table:yolo_detection_results}
\end{table}

\vspace*{-10pt}
\subsection{Object detection performance}

The optical microscopy images produced the strongest YOLOv26s detection performance, with mAP50 of 0.8865 and mAP50-95 of 0.6890. Among the holographic image modalities, p-type reconstructed holograms produced the best overall detection performance. At 100\% bounding box area, p-type reconstructed holograms achieved mAP50 of 0.8229 and mAP50-95 of 0.5324. This corresponds to 92.8\% of the optical mAP50 value and 77.3\% of the optical mAP50-95 value.

Raw holograms achieved mAP50 of 0.7694 and mAP50-95 of 0.4565 at 100\% bounding box area. R-type reconstructed holograms achieved mAP50 of 0.7528 and mAP50-95 of 0.4756 at 100\% bounding box area. These values show that the new holographic detection results are far above the old raw-hologram results on the previous dataset, where the best holographic detection value was 15.4\% mAP50 even after synthetic-data augmentation~\cite{warshaneyan2026sivp} (13.3\% before augmentation~\cite{warshaneyan2025ieee}). However, the optical baseline remained stronger, especially under the stricter mAP50-95 criterion.

\subsection{Object classification performance}

The MobileNetV4 variant was chosen on the raw-hologram 100\% condition: the Hybrid Medium variant was unstable, collapsing whenever more than its final backbone block was fine-tuned, and its best stable configuration (accuracy 0.5334, macro-F1 0.5306) stayed below Conv Medium (0.5731, 0.5832; Table~S5), so Conv Medium was used for all classification experiments.

Following this architecture screen, all remaining classification experiments used MobileNetV4 Conv Medium. Optical microscopy images produced the strongest classification performance, with accuracy of 0.9705 and macro-F1 of 0.9687. This confirms that optical classification remained close to the ceiling observed in the previous studies, where optical MobileNetV3L accuracy was approximately 97\%. The 100\% bounding-box-area MobileNetV4 Conv Medium classification results are shown in Table~\ref{table:mobilenet_classification_results}.

\begin{table}[!t]
\centering
\caption{\textbf{MobileNetV4 Conv Medium object classification performance on the test subset at 100\% bounding box area.} Expanded-area results are given in Table~S4.}
\begin{adjustbox}{max width=\linewidth}\begin{tabular}{|C{1.20in}|C{0.60in}|C{0.58in}|C{0.73in}|}
\hline
\textbf{Modality} & \textbf{Box area} & \textbf{Accuracy} & \textbf{Macro-F1}\\ \hline
Raw hologram & 100\% & 0.5731 & 0.5832\\ \hline
P-type hologram & 100\% & 0.6204 & 0.6153\\ \hline
R-type hologram & 100\% & 0.7866 & 0.7695\\ \hline
Optical microscopy & 100\% & 0.9705 & 0.9687\\ \hline
\end{tabular}\end{adjustbox}
\begin{flushleft}
Classification metrics are reported for the test subset using the best checkpoint.
\end{flushleft}
\label{table:mobilenet_classification_results}
\end{table}

Among the holographic image modalities, r-type reconstructed holograms produced the strongest classification performance.

At 100\% bounding box area, r-type reconstructed holograms achieved accuracy of 0.7866 and macro-F1 of 0.7695. This corresponds to 79.4\% of the optical macro-F1 value and 81.1\% of the optical accuracy value. Although this is not near-parity with optical microscopy, it represents a substantial improvement over the previous raw hologram classification result of 54\% accuracy.

Bounding-box expansion (125\% and 150\%; Table~S4) helped p-type crops consistently (macro-F1 0.6153 to 0.6590), helped raw holograms only at 150\% (0.5832 to 0.6044 via 0.5764), and hurt r-type crops (0.7695 to 0.7086). The reason is given in the interpretation below.

\subsection{Explainable AI analysis}

The explainability results are organized along the three evaluation axes: where the explanations concentrate, whether they are faithful, and how they behave across methods and reconstructions.

\subsection{Detection explanations}
Spatial grounding is strong on every modality on which it was measured (Table~\ref{table:xai_spatial}). Layer-CAM-G places 96.3\% of its map energy inside the manually annotated grain boxes on optical images, 98.6\% on p-type reconstructions, and 98.0\% on r-type reconstructions, corresponding to roughly 140-fold the geometric-chance level established by the random-map null. The remaining methods concentrate less sharply but all remain significantly above chance wherever applied, with localization lifts of 40 to 51$\times$ (Occlusion), 17 to 20$\times$ (IxG), and 1.2 to 1.4$\times$ (D-RISE and MFPP-D); the lifts hold or increase under holography (Fig~S6). Explanation faithfulness is summarized in Table~\ref{table:xai_detection}: the region-based methods retain 66 to 80\% of their Over-all score when moving from optical images to p-type reconstructions, whereas IxG collapses toward the random floor, a failure analyzed below. The deletion halves of these scores saturate because holographic models are brittle: deleting a random tenth of the pixels preserves about 64\% of an optical model's confidence but only 0.2\% of a p-type model's.

\begin{table*}[!t]
\centering
\caption{\textbf{Spatial grounding of detection explanations against manual ground truth.}
EBPG is the fraction of map energy inside the true box. Lift divides EBPG by the box share of
the image area (multiples of geometric chance). Sanity: pass means the map changes strongly
under model randomization.}
\begin{adjustbox}{max width=\textwidth}\begin{tabular}{|C{1.2in}|C{0.72in}|C{0.72in}|C{0.72in}|C{0.72in}|C{0.7in}|}
\hline
\textbf{Method} & \textbf{Optical EBPG} & \textbf{P-type EBPG} & \textbf{Optical lift} & \textbf{P-type lift} & \textbf{Sanity}\\ \hline
Layer-CAM-G & 0.9632 & 0.9861 & 141$\times$ & 145$\times$ & pass\\ \hline
Occlusion & 0.3108 & 0.3500 & 40$\times$ & 51$\times$ & pass\\ \hline
IxG & 0.1323 & 0.1367 & 17$\times$ & 20$\times$ & fail\\ \hline
MFPP-D & 0.0101 & 0.0100 & 1.4$\times$ & 1.4$\times$ & pass\\ \hline
D-RISE & 0.0084 & 0.0093 & 1.2$\times$ & 1.4$\times$ & partial\\ \hline
Random null & 0.0067 & 0.0067 & 1$\times$ & 1$\times$ & not applicable\\ \hline
\end{tabular}\end{adjustbox}
\begin{flushleft}
Layer-CAM-G on r-type reconstructions reaches 0.9801 (146$\times$). Gradient methods fail the
randomization check by construction of their known weaknesses; their spatial values are reported
for completeness.
\end{flushleft}
\label{table:xai_spatial}
\end{table*}

\begin{table*}[!t]
\centering
\caption{\textbf{Explanation faithfulness (Over-all score) by method and image modality.}
Higher is better; a random attribution map scores approximately zero on both modalities.}
\begin{adjustbox}{max width=\textwidth}\begin{tabular}{|C{1.7in}|C{1.15in}|C{0.72in}|C{0.72in}|C{0.72in}|}
\hline
\textbf{Method} & \textbf{Family} & \textbf{Optical} & \textbf{P-type} & \textbf{Retained}\\ \hline
Occlusion & perturbation & 0.7341 & 0.5859 & 79.8\%\\ \hline
MFPP-D & perturbation & 0.583 & 0.469 & 80.4\%\\ \hline
D-RISE & perturbation & 0.641 & 0.443 & 69.1\%\\ \hline
Layer-CAM-G & activation & 0.6567 & 0.4347 & 66.2\%\\ \hline
IxG & gradient & 0.5759 & 0.0602 & 10.5\%\\ \hline
Random baseline & control & +0.003 & +0.000 & not applicable\\ \hline
\end{tabular}\end{adjustbox}
\begin{flushleft}
Occlusion and IxG cover all matched detections (381 optical, 313 p-type); Layer-CAM-G faithfulness uses an unbiased random subsample (194 and 161), and D-RISE and MFPP-D values are from the stored per-detection result files of the original full evaluation runs. The deletion half of the score is saturated by model brittleness on both modalities, so the retained fractions are driven by the insertion half. A full-coverage rerun of Layer-CAM-G under a stricter end-to-end scoring protocol (not comparable in absolute value) confirms the ordering and adds the r-type column: retention 62.7\% for p-type and 58.4\% for r-type, mirroring the detection-accuracy ordering of the two reconstructions.
\end{flushleft}
\label{table:xai_detection}
\end{table*}

\begin{figure*}[!t]
\centering
\begin{minipage}{0.46\textwidth}\centering
\includegraphics[width=\linewidth]{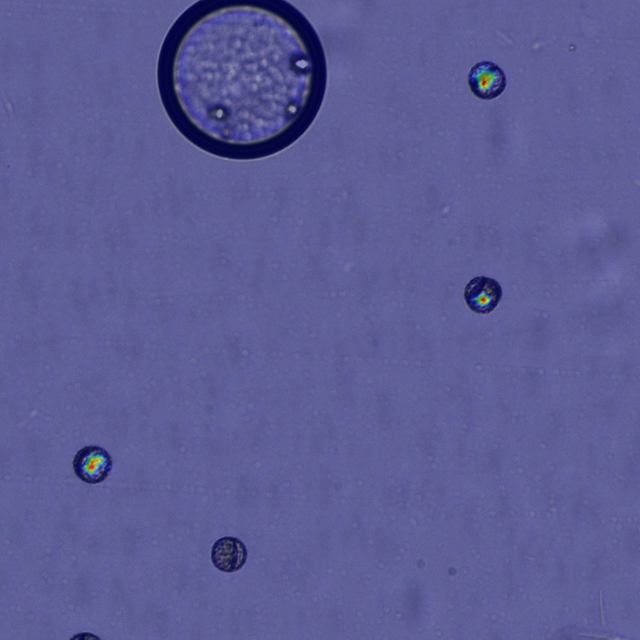}\\[3pt]
\small (a) Optical patch, Layer-CAM-G
\end{minipage}\hspace{0.04\textwidth}
\begin{minipage}{0.46\textwidth}\centering
\includegraphics[width=\linewidth]{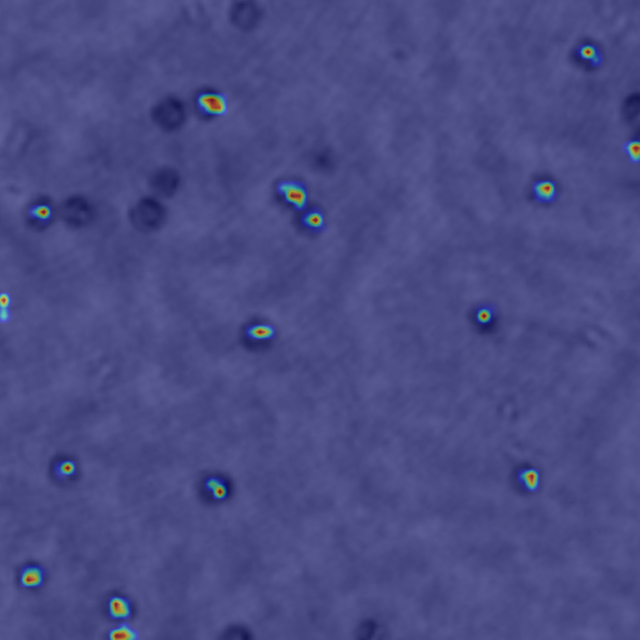}\\[3pt]
\small (b) P-type hologram patch, Layer-CAM-G
\end{minipage}
\caption{\textbf{Representative Layer-CAM-G saliency maps, optical versus p-type.} Layer-CAM-G attribution overlaid on (a) an optical patch and (b) a p-type reconstruction patch for the same species. Maps are computed per detection within 640-pixel patches, and the attribution concentrates on the target grain in both modalities.}
\label{fig:xai_saliency}
\end{figure*}

\subsection{Classification explanations}
The classification-side pattern differs from the detection side (Table~\ref{table:xai_classification}). All values are computed at the classifier training resolution of 256 pixels with each modality's own checkpoint and a paired random map per crop. On optical crops every method exceeds its random floor (Layer-CAM +0.127, 180 of 252 crops; Occlusion +0.070, 148 of 252; IG +0.034; IxG +0.009). On the holographic crops the methods split: perturbation-based Occlusion remains strongly faithful (r-type +0.140 over random, 199 of 248 crops, sign-test $p$~=~1.3$\times$10$^{-22}$, Cohen's $d$~=~0.86; p-type +0.177, 220 of 248, $p$~=~3.8$\times$10$^{-38}$, $d$~=~1.22), whereas the gradient- and activation-based methods fall to or below their random floor on r-type crops (Layer-CAM $-$0.020, IG $-$0.009, IxG $-$0.015) and, except Layer-CAM (+0.044, 154 of 248), on p-type crops as well. Perturbation-based explanation therefore transfers to holographic classification where gradient-based explanation does not, and it is the method of choice for that arm. The failure of gradient attribution on r-type crops is not repaired by smoothing either (next subsection).

\begin{table}[!t]
\centering
\caption{\textbf{Classification explanation faithfulness (Over-all score) by method and modality.} Higher is better; paired random baselines score near zero throughout. All values at the classifier training resolution of 256 pixels with the modality's own checkpoint (252 optical, 248 r-type, 248 p-type crops).}
\begin{adjustbox}{max width=\linewidth}\begin{tabular}{|C{1.4in}|C{1.05in}|C{1.05in}|C{1.05in}|}
\hline
\textbf{Method} & \textbf{Optical} & \textbf{R-type} & \textbf{P-type}\\ \hline
Layer-CAM & +0.1292 & $-$0.0215 & +0.0462\\ \hline
Integrated Gradients & +0.0361 & $-$0.0108 & $-$0.0013\\ \hline
Occlusion & +0.0726 & +0.1378 & +0.1792\\ \hline
IxG & +0.0115 & $-$0.0165 & $-$0.0061\\ \hline
Random baseline & +0.0022 & $-$0.0018 & +0.0023\\ \hline
\end{tabular}\end{adjustbox}
\label{table:xai_classification}
\end{table}

\subsection{A granularity mechanism behind gradient-explanation collapse}
\begin{figure*}[!t]
\centering
\begin{minipage}{0.31\textwidth}\centering
\includegraphics[width=\linewidth]{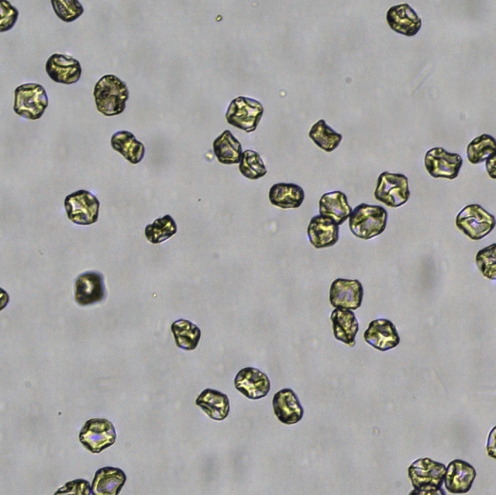}\\[3pt]
\small (a) Patch
\end{minipage}\hspace{0.02\textwidth}
\begin{minipage}{0.31\textwidth}\centering
\includegraphics[width=\linewidth]{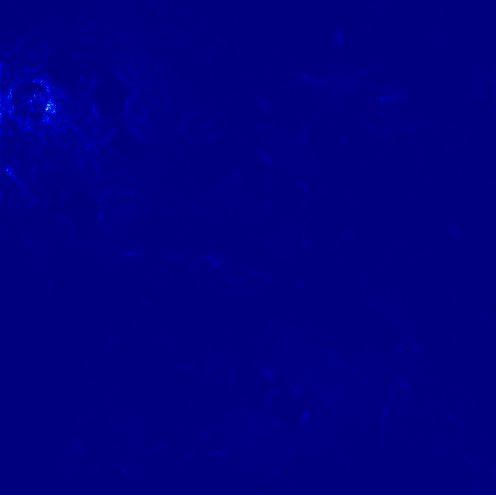}\\[3pt]
\small (b) Plain Input $\times$ Gradient map
\end{minipage}\hspace{0.02\textwidth}
\begin{minipage}{0.31\textwidth}\centering
\includegraphics[width=\linewidth]{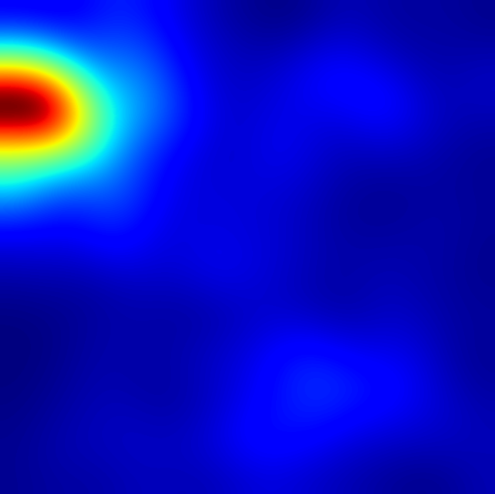}\\[3pt]
\small (c) After spatial smoothing
\end{minipage}
\caption{\textbf{IxG attribution maps before and after spatial smoothing.} (a) An image patch, (b) its pixel-scattered plain Input $\times$ Gradient map, and (c) the same map after spatial smoothing. Smoothing turns a correct but fragmented ranking into region-scale structure.}
\label{fig:xai_blurmaps}
\end{figure*}

\begin{figure*}[!t]
\centering
\includegraphics[width=0.9\textwidth]{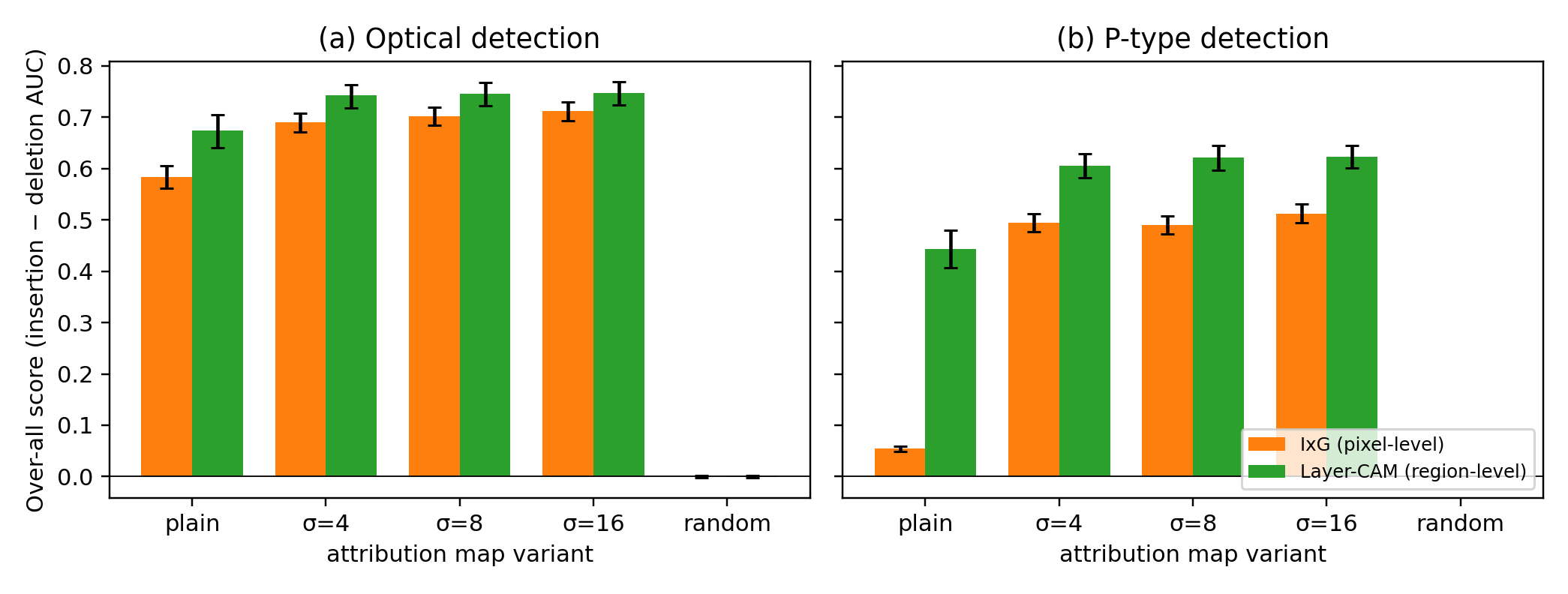}
\caption{\textbf{Granularity rescue is method-specific.} Over-all score of the plain, smoothed ($\sigma$~=~4, 8, 16) and random variants of pixel-level IxG maps and region-level plain Layer-CAM maps on the same matched detections, (a) optical and (b) p-type; error bars are 95\% bootstrap confidence intervals.}
\label{fig:rescue}
\end{figure*}

The collapse of IxG is not a misattribution. Its spatial grounding is stable and 17 to 20$\times$ above chance on both detection modalities, so the maps point at the correct grains. What fails is the \emph{texture} of the ranking, in which importance is scattered over isolated pixels that never assemble into coherent structures during region-based scoring. The decisive test is to smooth the finished attribution map spatially, an operation that adds no model information whatsoever (Table~\ref{table:xai_blur}, Fig~\ref{fig:xai_blurmaps}). On detection, smoothing at $\sigma$~=~16 raises the p-type Over-all score from +0.054 to +0.512 [+0.494, +0.530], with all 306 detections improving (sign-test $p$~=~1.5$\times$10$^{-92}$), closing about 88\% of the gap to the optical value, which itself rises from +0.584 to +0.712. Smoothing helps region-level maps far less: on the same detections, plain Layer-CAM gains +0.180 on p-type against +0.458 for IxG (paired difference +0.278 [0.244, 0.311], IxG larger on 258 of 306 detections, bootstrap probability 1.000), so the benefit scales with map granularity (Fig~\ref{fig:rescue}). The gain is not driven by the most concentrated maps: per-item map concentration (Gini index) correlates negatively with the smoothing gain (Spearman $\rho$~=~$-$0.375 optical, $-$0.696 p-type; partial correlations $-$0.392 and $-$0.712 controlling for the plain score), indicating that smoothing consolidates fragmented but present regional mass, and that the most pixel-sparse maps are rescued least.
 On classification, smoothing raises the optical IxG score from +0.014 to +0.126, approaching Layer-CAM, whereas on holographic crops it does not help: r-type moves from $-$0.015 to $-$0.008 and p-type from $-$0.006 to +0.005, both with the confidence interval of the gain covering zero.
 The effect is robust across a threefold range of smoothing scales, whereas five pre-declared rescue attempts operating on the attribution values themselves (SmoothGrad and related estimators~\cite{smilkov2017smoothgrad}) all failed, confirming that the failure is spatial rather than statistical.

\begin{table}[!t]
\centering
\caption{\textbf{Effect of spatial map smoothing on IxG faithfulness (plain $\rightarrow$ smoothed).}}
\begin{adjustbox}{max width=\linewidth}\begin{tabular}{|C{1.5in}|C{1.25in}|C{1.25in}|C{1.35in}|}
\hline
\textbf{Task} & \textbf{Optical} & \textbf{R-type} & \textbf{P-type}\\ \hline
Detection & +0.584 $\rightarrow$ +0.712 & not evaluated & +0.054 $\rightarrow$ +0.512\\ \hline
Classification & +0.014 $\rightarrow$ +0.126 & $-$0.015 $\rightarrow$ $-$0.008 & $-$0.006 $\rightarrow$ +0.005\\ \hline

\end{tabular}\end{adjustbox}
\begin{flushleft}
Detection values from the per-item smoothing experiment (372 optical and 306 p-type matched detections), not comparable in absolute value with Table~\ref{table:xai_detection}, and r-type detection lies outside the perturbation-method design pattern. Classification values at 256 pixels with each modality\textquotesingle s own checkpoint. Smoothed IxG approaches the Layer-CAM level on optical crops only, and on both holographic classification arms the gain is within the confidence interval of zero.

\end{flushleft}
\label{table:xai_blur}
\end{table}

\subsection{Cross-method agreement and reconstruction invariance}
\begin{figure*}[!t]
\centering
\includegraphics[width=0.9\textwidth]{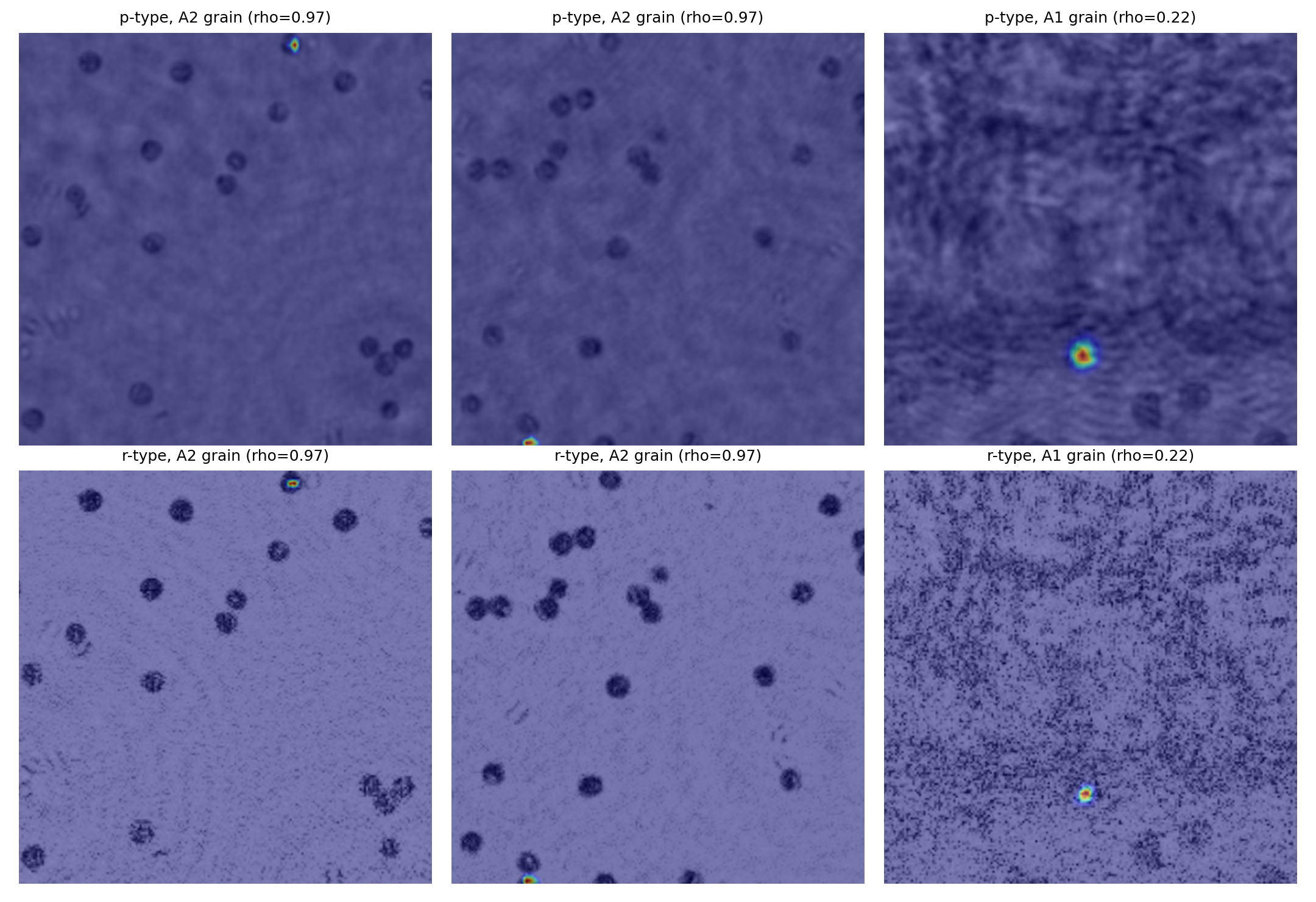}
\caption{\textbf{Matched-grain saliency maps for the two reconstructions.}
Layer-CAM-G attribution overlaid on the p-type (top row) and r-type (bottom row) patches
for the same physical grains detected at identical coordinates; two high-agreement pairs and one
low-agreement pair are shown with their map rank correlations. Panels are full 640-pixel patches.}
\label{fig:xai_samegrain}
\end{figure*}

The attribution methods agree with one another well above chance on which detections are hard to explain (Kendall's $W$~=~0.55 on optical and 0.48 on p-type images), and this agreement is statistically indistinguishable between the modalities once the analysis resamples whole patches rather than individual detections. The methods also jointly flag the least-explainable holographic detections far above chance ($p$~=~1.6$\times$10$^{-5}$). Finally, the two reconstructions produce closely matching explanations for the same physical grains: for grains detected by both pipelines at identical coordinates, the saliency maps agree at a median rank correlation of 0.83, the in-box energy fractions do not differ significantly, and the per-species difficulty profiles of the two reconstructions correlate at 0.94. For grains detected by both pipelines, the explanations are therefore robust to the choice of reconstruction algorithm, and Fig~\ref{fig:xai_samegrain} illustrates a matched pair.

\subsection{Noise gate, dose-response and validation matrix}
\begin{figure*}[!t]
\centering
\includegraphics[width=0.9\textwidth]{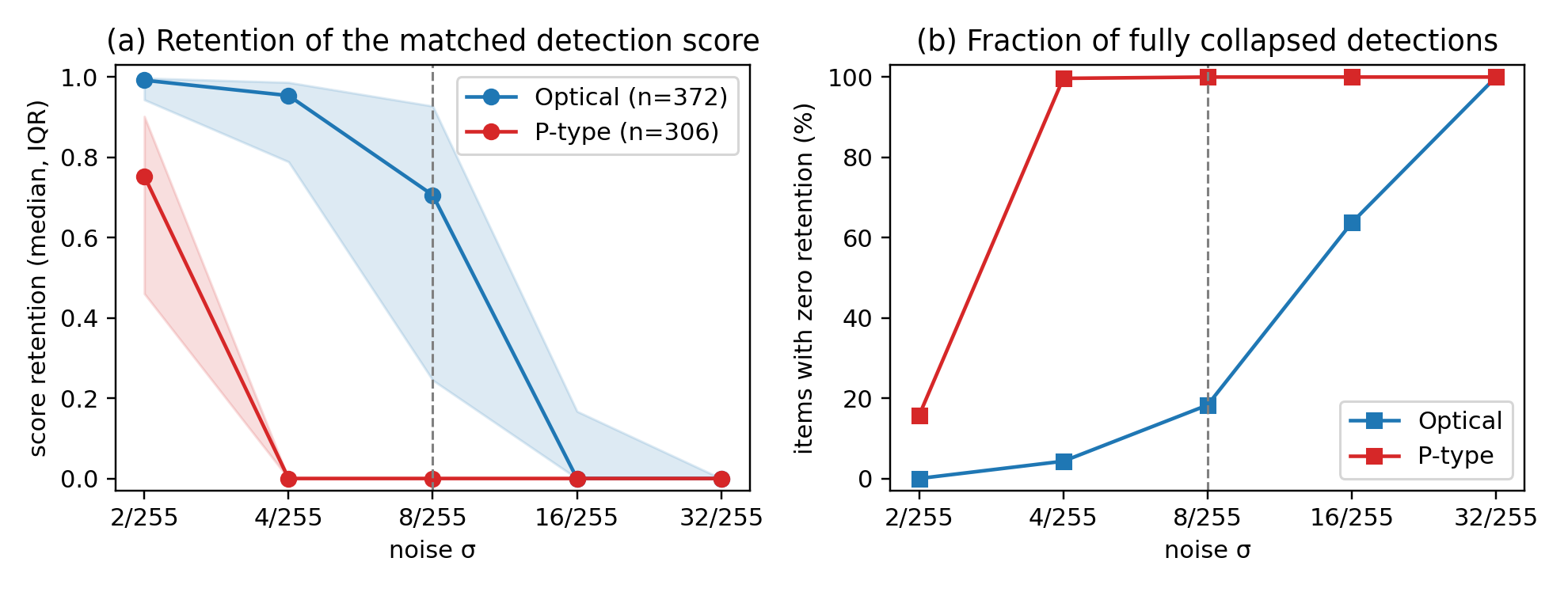}
\caption{\textbf{Noise-gate dose-response.} (a) Median retention of the matched detection score with interquartile range and (b) fraction of fully collapsed detections as a function of the noise standard deviation; the dashed line marks the operating point of 8/255.}
\label{fig:gate_dose}
\end{figure*}

Table~\ref{table:gate} applies the AHIR noise gate to the seeded per-arm gate cohorts (noise $\sigma$~=~8/255, five draws). Optical detections retain a median 0.702 [0.612, 0.794] of their score, with 18.8\% of items collapsing to zero, whereas p-type detections retain nothing at all: 306 of 306 items collapse, although their clean confidences are healthy (per-species means 0.33 to 0.78, overlapping the optical range), so the collapse is intrinsic brittleness of the holographic detector rather than weak detection. Its deletion-based faithfulness is therefore declared not interpretable. On classification the optical arm is the most brittle passing arm (median retention 0.024; 70\% of crops below 0.5) despite near-ceiling accuracy, whereas p-type crops are predominantly brittle (median 0.363, 60\% below 0.5) and r-type crops are heterogeneous (median 0.625 [0.357, 0.822]) in a way that follows the classifier's per-species accuracy. The gate is validated on the optical detection arm: gate-passing detections reach an Over-all score of +0.674 [+0.651, +0.696] ($n$~=~228) against +0.442 [+0.409, +0.476] ($n$~=~144) for gate-failing ones, non-overlapping intervals, so the gate stratifies faithfulness as intended. Brittleness is a property of species features rather than of confidence (at an identical clean probability, A1 crops retain their full score while A2 to A5 crops collapse; Section~S6), which is why the gate must operate per item. The dose-response sweep (Fig~\ref{fig:gate_dose}) derives the operating point: optical retention falls from 0.99 at $\sigma$~=~2/255 through 0.95 and 0.71 to complete collapse at 16/255, whereas p-type detection is at 0.75 at 2/255 and fully collapsed from 4/255 onward, a four-fold lower failure threshold. The value $\sigma$~=~8/255 is the largest value at which optical retains majority confidence while p-type has already collapsed completely.

\begin{table*}[!t]
\centering
\caption{\textbf{Noise-gate matrix.} Median score retention under pixel noise ($\sigma$~=~8/255, five draws) with 95\% bootstrap intervals, fraction of items with zero retention, fraction below 0.5, and the resulting verdict on deletion-based faithfulness.}
\begin{adjustbox}{max width=\textwidth}\begin{tabular}{|C{1.15in}|C{0.45in}|C{1.35in}|C{0.7in}|C{0.7in}|C{1.1in}|}
\hline
\textbf{Arm} & \textbf{n} & \textbf{Median retention} & \textbf{Zero} & \textbf{Below 0.5} & \textbf{Verdict}\\ \hline
Detection, optical & 372 & 0.702 [0.612, 0.794] & 18.8\% & 38.7\% & interpretable\\ \hline
Detection, p-type & 306 & 0.000 [0.000, 0.000] & 100\% & 100\% & not interpretable\\ \hline
Classification, optical & 290 & 0.024 [0.021, 0.031] & 0.0\% & 70\% & interpretable, brittle\\ \hline
Classification, p-type & 248 & 0.363 [0.309, 0.421] & 0.0\% & 60.1\% & interpretable, brittle\\ \hline
Classification, r-type & 283 & 0.625 [0.357, 0.822] & 0.0\% & 47.7\% & interpretable, heterogeneous\\ \hline
\end{tabular}\end{adjustbox}
\label{table:gate}
\end{table*}

\subsection{Replication, decomposition of the localization score, and method generality}
The IxG faithfulness collapse replicates on an independent item basis: +0.584 [+0.562, +0.606] on optical ($n$~=~372) against +0.054 [+0.049, +0.059] on p-type ($n$~=~306), a retention of 9.3\% that matches the 10.5\% of Table~\ref{table:xai_detection}. The near-perfect localization of Layer-CAM-G is not conferred by its Gaussian term: plain Layer-CAM without that term reaches an in-box energy of 0.966 on optical and 0.991 on p-type detections (678 detections), matching the published values, whereas a Gaussian centred on the predicted box alone reaches 0.730 and 0.705 and a box-indicator map 0.897 and 0.838 (Table~S6). The score is earned by the method. Generality across methods holds on both tasks: G-CAM-E exceeds its own random floor by +0.663 [+0.517, +0.789] on optical (21 of 22 paired detections) and +0.381 [+0.238, +0.520] on p-type (21 of 24), and Occlusion exceeds its floor on every classification arm (Table~\ref{table:xai_classification}).

\subsection{Deletion- and insertion-side detector metrics}
\begin{figure*}[!t]
\centering
\includegraphics[width=0.9\textwidth]{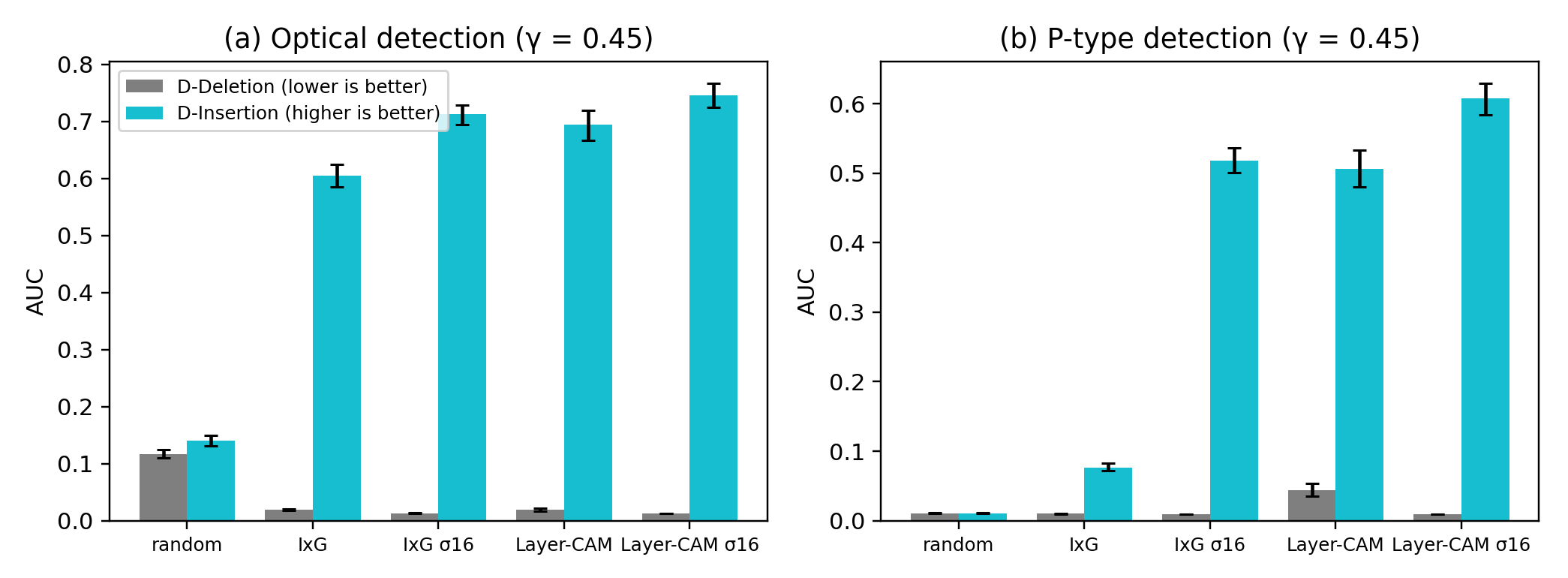}
\caption{\textbf{Deletion- versus insertion-side detector metrics.} D-Deletion and D-Insertion ($\gamma$~=~0.45) for random, plain and smoothed maps on (a) optical and (b) p-type detections; error bars are 95\% bootstrap intervals.}
\label{fig:dfamily}
\end{figure*}

Expressed in the published D-Deletion metric, optical detections separate cleanly from random (IxG 0.019 and plain Layer-CAM 0.019 against a random floor of 0.117; lower is better), but on p-type every map converges to the same floor of about 0.01, because the brittle detector disappears after roughly 3\% of the pixels are removed regardless of their ranking, and Layer-CAM even scores slightly worse than random there, and smoothing the maps moves the metric by $-$0.0006 for IxG, statistically real and practically nothing. The insertion-side counterpart stays informative on the same detections (Fig~\ref{fig:dfamily}, Table~S7): D-Insertion separates every map from random on p-type (IxG 0.077 against 0.011, 303 of 303 detections, $p$~=~1.2$\times$10$^{-91}$; plain Layer-CAM 0.507; smoothed IxG 0.518; smoothed Layer-CAM 0.607), and smoothing lifts IxG by +0.441 [+0.424, +0.459], 4.4 times the Layer-CAM gain, reproducing the granularity rescue inside the published metric family. A seeded subsample of 50 detections per modality computed on a separate CPU run reproduced these values within a few hundredths. The two members of the family thus fail and succeed for the reason the gate predicts: deletion measures brittleness on a gate-failing model, insertion measures attribution quality.

\subsection{Interpretation and limitations}

The results show that the newer dataset substantially reduced the gap between optical microscopy and holographic microscopy, but did not eliminate it. The earlier studies found that raw holograms produced weak object detection and classification performance compared with optical microscopy. In the present study, even the raw hologram baseline improved markedly over the previous raw hologram results. More importantly, reconstructed holographic images produced stronger results than raw holograms for the tasks they were best suited for.

The crop-area responses reflect what each reconstruction contains. The p-type image balances added information against added noise, so each successive expansion still adds usable context. The raw hologram takes a noise hit at the first expansion before the second adds enough information to pay for it, giving the observed dip and recovery. The r-type reconstruction is effectively a greyscale optical-like image: it carries no residual holographic information and no colour, so expansion adds background noise only.

The present manuscript reports the stricter macro-F1 and mAP50-95 as primary metrics, with accuracy and mAP50 retained for continuity with the previous studies, because the work has moved from a feasibility stage to a parity-oriented evaluation in which class-specific weakness and imperfect localization must not be hidden.

The combined detect-and-classify experiment showed why the two tasks were separated: missed detections receive no class prediction, so matched-detection classification metrics overstate what a single model delivers.

The explainability analysis adds a trust dimension to the modality comparison. Explanations remain functional under holography: every evaluated method stays noticeably above geometric chance in localization, the region-based methods keep the large majority of their faithfulness, and the amount retained follows the same ordering as raw task performance, with the reconstruction ranking of the explanations mirroring the accuracy ranking of the models. Two general observations emerge that extend beyond pollen imagery. First, holographic models are brittle in a way optical models are not: deleting a random tenth of the pixels preserves about 64\% of an optical model's confidence but only 0.2\% of a p-type model's, which systematically caps every perturbation-based explanation score and argues for decomposing deletion- and insertion-based measurements whenever brittle models are evaluated. Second, gradient-attribution methods can fail for reasons of map granularity rather than misattribution, and a parameter-light spatial smoothing of the finished map separates the two failure modes, and we suggest this as a routine diagnostic before concluding that a gradient method fails on a non-natural imaging domain. The two findings are two distinct failure modes with two distinct remedies: score brittleness, which the noise gate detects and which no map operation can repair, and ranking granularity, which the smoothing correction repairs. The published deletion-side detector metric fails under the first mode while its insertion-side counterpart does not.

Several limitations bound these conclusions. The brittleness ceiling means the absolute faithfulness values on holograms understate explanation quality to an unknown degree. The smoothing repair has one boundary condition, r-type classification, where it does not help. Sanity checks were passed convincingly by the activation-based methods but only partially by the perturbation methods, and the gradient methods inherit known weaknesses in this respect~\cite{adebayo2018sanity}. Finally, the detector-specific D-RISE and MFPP-D analyses were scoped to the primary optical-versus-p-type detection comparison, where p-type was the strongest holographic detector. The r-type arm was retained as a secondary reconstruction check with Layer-CAM-G, and conclusions involving D-RISE and MFPP-D therefore apply to the optical and p-type detection arms.

\section{Conclusions}

This study shows that improved holographic image preparation substantially increases deep learning performance for automated pollen analysis: optical microscopy remains the strongest modality, but the gap to the reconstructed holographic variants has narrowed considerably compared with the previous dataset.

P-type reconstructions are the strongest holographic modality for detection and r-type reconstructions for classification, and the architecture screen supports the convolutional MobileNetV4 variant, and parity with optical microscopy is not yet reached, but the progress toward it is substantial.

The results also show that a single YOLO model trained to perform detection and classification together is not sufficient for this dataset. The combined model produced misleadingly high classification metrics when evaluated only on matched detections, but its all-object coverage-adjusted classification performance was substantially lower. This supports a two-model pipeline in which YOLO performs object detection and MobileNet performs crop-level object classification.

Overall, the current dataset and reconstruction-based image preparation produced much stronger holographic performance than the earlier raw-hologram studies. Future work should focus on improving strict localization performance for holographic object detection, increasing class-balanced classification performance for reconstructed holograms and exploring reconstruction or training strategies that reduce the remaining gap with optical microscopy.

Beyond raw performance, the explainability analysis shows that the holographic models' decisions are grounded where they should be: activation-based saliency methods concentrate on the true grains wherever applied, region-based methods retain most of their faithfulness under holography, and produce closely matching explanations across the two reconstructions. The analysis also yields two transferable findings, the pronounced perturbation brittleness of holographic models and a granularity mechanism that explains and repairs the apparent failure of gradient-based explanations, and it packages them as the reusable AHIR protocol: gate first, then correct. Together these results indicate that reconstruction-based DIHM pipelines can be made not only more accurate but also more trustworthy, which is a prerequisite for eventual veterinary deployment.

\section*{Supplementary material}
Supplementary material associated with this article is included at the end of this preprint.

\section*{Data availability}
Summary results are contained within the paper and its supplementary material. Additional intermediate and final experiment and computation results underlying the reported metrics may be available from the corresponding author on reasonable request for verification. The raw microscopy images, holograms, reconstructions, and annotation labels are not publicly available.

\section*{Funding}
This research was funded by the Latvian Council of Science, project VetCyto, Nr. lzp-2023/1-0220.

\section*{Declaration of competing interest}
The authors declare no competing interests.

\section*{Ethics approval}
No ethics approval was required for this project.

\section*{Declaration of generative AI and AI-assisted technologies in the manuscript preparation process}
During the preparation of this work, the authors used Claude and ChatGPT to assist with selected text revisions for clarity and quality. The authors reviewed and edited the outputs as needed and take full responsibility for the content of the publication.

\bibliographystyle{elsarticle-num}
\bibliography{refs}

\clearpage
\onecolumn
\setcounter{table}{0}
\setcounter{figure}{0}
\begingroup
\renewcommand{\thefootnote}{\fnsymbol{footnote}}
\begin{center}
\Large Supplementary material for: Reconstructed holograms and explanation-aware evaluation for low-cost computational pollen analysis in veterinary cytology\par\vskip18pt
\normalsize Swarn Warshaneyan\textsuperscript{a,}\footnotemark[1], Joial Danyal\textsuperscript{a}, Bla\v{z} Cugmas\textsuperscript{b}, Mindaugas Tamo\v{s}i\={u}nas\textsuperscript{b}, Edgars Kviesis-Kipge\textsuperscript{b}, Kirishanth Manivannan\textsuperscript{a}, Roberts Kadi\c{k}is\textsuperscript{a}\par\vskip10pt
\footnotesize\itshape \textsuperscript{a}Institute of Electronics and Computer Science (EDI), Riga, Latvia\par
\textsuperscript{b}Faculty of Science and Technology, University of Latvia, Riga, Latvia\par\vskip36pt
\hrule\vskip12pt
\hrule\vskip12pt
\end{center}
\footnotetext[1]{Corresponding author.}
\renewcommand{\thefootnote}{}
\footnotetext{Email address: swarn.warshaneyan@gmail.com (Swarn Warshaneyan)}
\endgroup
\renewcommand{\thetable}{S\arabic{table}}
\renewcommand{\thefigure}{S\arabic{figure}}
\renewcommand{\theHtable}{supp.\arabic{table}}
\renewcommand{\theHfigure}{supp.\arabic{figure}}
\section*{Supplementary tables and figures}
\begin{table}[H]
\centering
\caption{\textbf{Per-class object-instance counts before and after blackening removal.} Pre-crop
counts equal the optical label counts; the post-crop count is shared by the raw, p-type, and r-type
holographic modalities.}
\begin{adjustbox}{max width=\linewidth}\begin{tabular}{|C{0.9in}|C{1.2in}|C{1.3in}|C{1.0in}|}
\hline
\textbf{Class} & \textbf{Optical (pre-crop)} & \textbf{Hologram (post-crop)} & \textbf{Retained (\%)}\\ \hline
A1 & 30285 & 20553 & 67.87\\ \hline
A2 & 34159 & 23992 & 70.24\\ \hline
A3 & 39570 & 28931 & 73.11\\ \hline
A4 & 51782 & 36758 & 70.99\\ \hline
A5 & 16980 & 12083 & 71.16\\ \hline
A7 & 54352 & 39438 & 72.56\\ \hline
\textbf{Total} & \textbf{227128} & \textbf{161755} & \textbf{71.22}\\ \hline
\end{tabular}\end{adjustbox}
\label{table:instances}
\end{table}

\begin{table}[H]
\centering
\caption{\textbf{Train/validation/test split.} Slide-level 8:1:1 split; instance counts shown for
optical (pre-crop) and holographic (post-crop).}
\begin{adjustbox}{max width=\linewidth}\begin{tabular}{|C{1.0in}|C{0.8in}|C{1.2in}|C{1.2in}|}
\hline
\textbf{Split} & \textbf{Images} & \textbf{Optical instances} & \textbf{Hologram instances}\\ \hline
Train & 48 & 163393 & 114289\\ \hline
Validation & 6 & 24885 & 17875\\ \hline
Test & 6 & 38850 & 29591\\ \hline
\textbf{Total} & \textbf{60} & \textbf{227128} & \textbf{161755}\\ \hline
\end{tabular}\end{adjustbox}
\label{table:splits}
\end{table}

\begin{table}[H]
\centering
\caption{\textbf{Single-model YOLO detect-and-classify experiment on raw holograms at 100\% bounding box area.}}
\begin{adjustbox}{max width=\linewidth}\begin{tabular}{|C{1.35in}|C{1.55in}|C{1.85in}|}
\hline
\textbf{Metric} & \textbf{Matched detections only} & \textbf{Coverage-adjusted all-object value}\\ \hline
Accuracy & 0.8334 & 0.4268\\ \hline
Macro-F1 & 0.6492 & 0.3324\\ \hline
\end{tabular}\end{adjustbox}
\begin{flushleft}
The matched-detection classification metrics use only objects that were successfully detected and matched. The coverage-adjusted values count missed ground-truth objects as classification failures, making them more comparable with crop-based MobileNet classification metrics. This experiment used 18,225 matched detections from 35,591 ground-truth objects, giving a matched-object coverage of 0.5121.
\end{flushleft}
\label{table:yolo_combined_task}
\end{table}

\begin{table}[H]
\centering
\caption{\textbf{MobileNetV4 Conv Medium object classification performance on the test subset using best checkpoints.}}
\begin{adjustbox}{max width=\linewidth}\begin{tabular}{|C{1.20in}|C{0.60in}|C{0.58in}|C{0.73in}|}
\hline
\tableheader{Image}{modality} & \tableheader{Bounding}{box area} & \textbf{Accuracy} & \textbf{Macro-F1}\\ \hline
Raw hologram & 100\% & 0.5731 & 0.5832\\ \hline
Raw hologram & 125\% & 0.5698 & 0.5764\\ \hline
Raw hologram & 150\% & 0.5942 & 0.6044\\ \hline
\tablecell{P-type\\reconstructed\\hologram} & 100\% & 0.6204 & 0.6153\\ \hline
\tablecell{P-type\\reconstructed\\hologram} & 125\% & 0.6498 & 0.6512\\ \hline
\tablecell{P-type\\reconstructed\\hologram} & 150\% & 0.6577 & 0.6590\\ \hline
\tablecell{R-type\\reconstructed\\hologram} & 100\% & 0.7866 & 0.7695\\ \hline
\tablecell{R-type\\reconstructed\\hologram} & 125\% & 0.7477 & 0.7279\\ \hline
\tablecell{R-type\\reconstructed\\hologram} & 150\% & 0.7464 & 0.7086\\ \hline
Optical microscopy & 100\% & 0.9705 & 0.9687\\ \hline
\end{tabular}\end{adjustbox}
\begin{flushleft}
Classification metrics are reported for the test subset using the best checkpoint.
\end{flushleft}
\label{table:S_mobilenet_full}
\end{table}

\begin{table}[H]
\centering
\caption{\textbf{MobileNetV4 architecture screening on raw holograms at 100\% bounding box area.}}
\label{table:mobilenet_architecture_screen}
\begin{adjustbox}{max width=\linewidth}\begin{tabular}{|C{1.20in}|C{1.35in}|C{0.70in}|C{0.70in}|}
\hline
\textbf{Architecture} & \textbf{Phase 2 fine-tuning} & \textbf{Accuracy} & \textbf{Macro-F1}\\ \hline
MobileNetV4 Hybrid Medium & Final top-level backbone block & 0.5334 & 0.5306\\ \hline
MobileNetV4 Conv Medium & Full backbone & 0.5731 & 0.5832\\ \hline
\end{tabular}\end{adjustbox}
\begin{flushleft}
Metrics are reported for the test subset using the best checkpoint. The Hybrid Medium row represents the best-performing tested hybrid configuration.
 Deeper Phase 2 unfreezing degraded the Hybrid variant monotonically: test accuracy 0.53 with one block unfrozen, 0.44 with two blocks, and 0.44 with all five blocks.\end{flushleft}
\end{table}

\begin{figure}[H]
\centering
\includegraphics[width=0.85\linewidth]{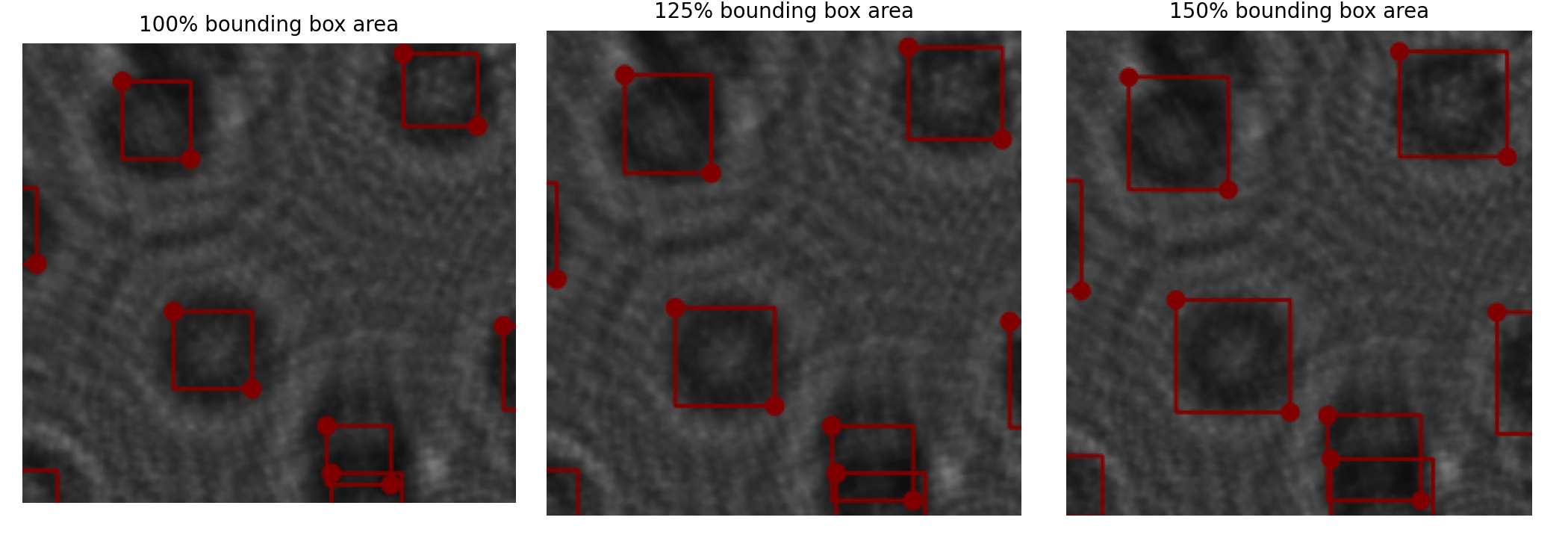}
\caption{\textbf{Bounding box areas of 100\%, 125\%, and 150\% on a raw hologram.}}
\label{fig:expansion}
\end{figure}

\begin{figure}[H]
\centering
\includegraphics[width=0.85\linewidth]{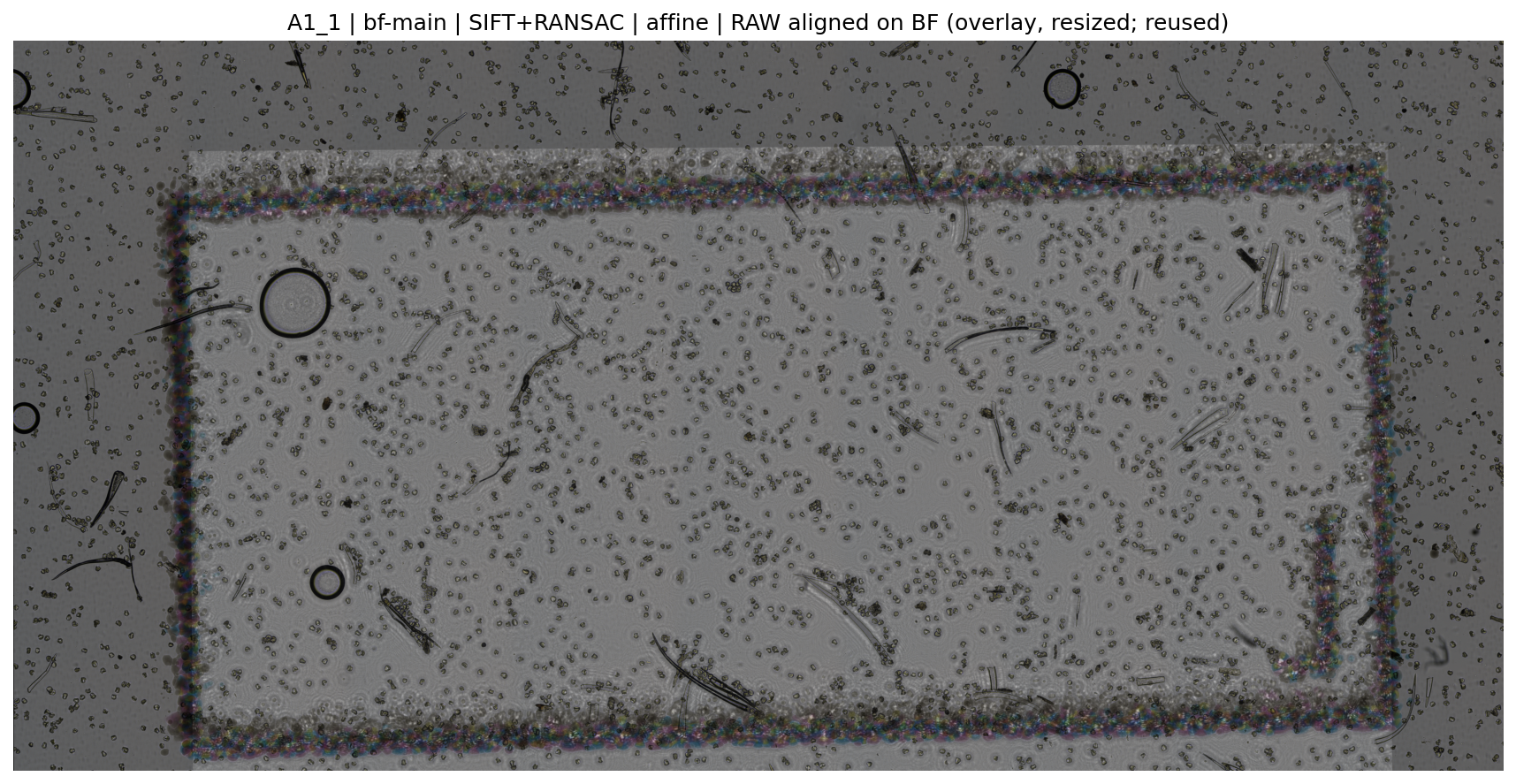}
\caption{\textbf{Size gap between the modalities after alignment.}
The raw hologram aligned onto the brightfield reference; the surrounding margin shows how much
smaller the holographic field of view is. Warped regions falling outside the holographic frame
become black borders, and crops containing them were removed.}
\label{fig:blackening}
\end{figure}

\begin{figure}[H]
\centering
\includegraphics[width=0.85\linewidth]{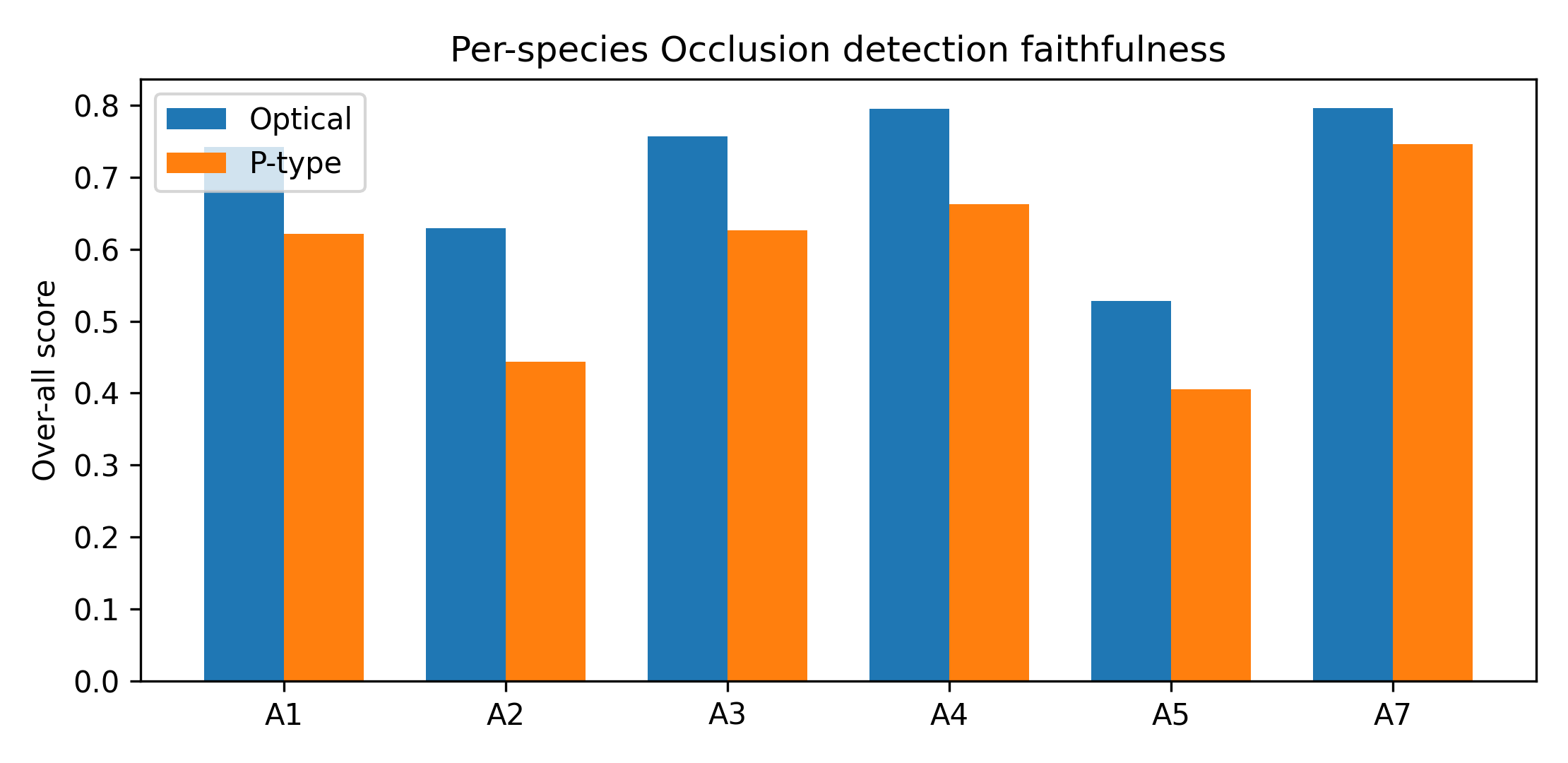}
\caption{\textbf{Per-species Occlusion detection explanation faithfulness.} Over-all score per pollen species for optical and p-type images.}
\end{figure}
\begin{figure}[H]
\centering
\includegraphics[width=0.85\linewidth]{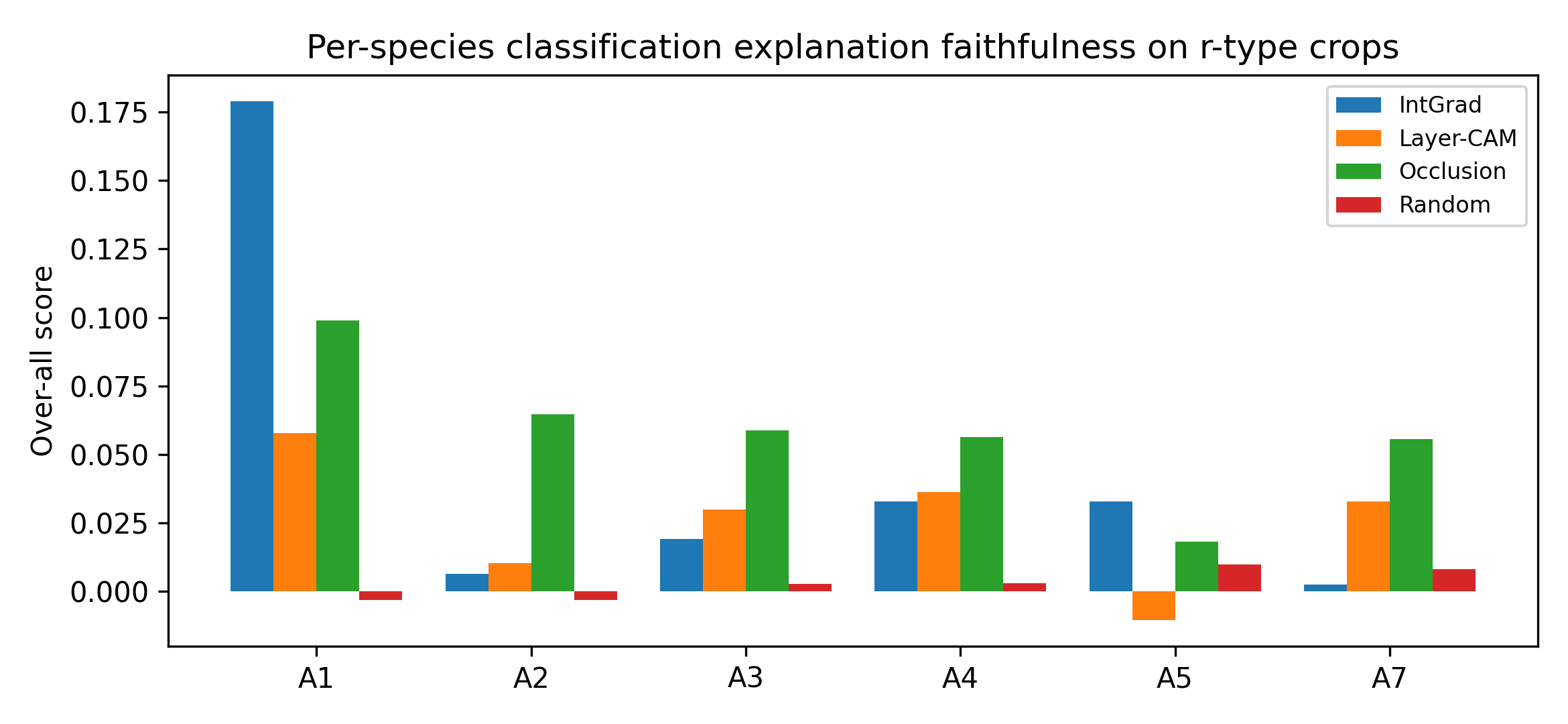}
\caption{\textbf{Per-species classification explanation faithfulness on r-type crops.} Over-all score per pollen species and attribution method.}
\end{figure}
\begin{figure}[H]
\centering
\includegraphics[width=0.85\linewidth]{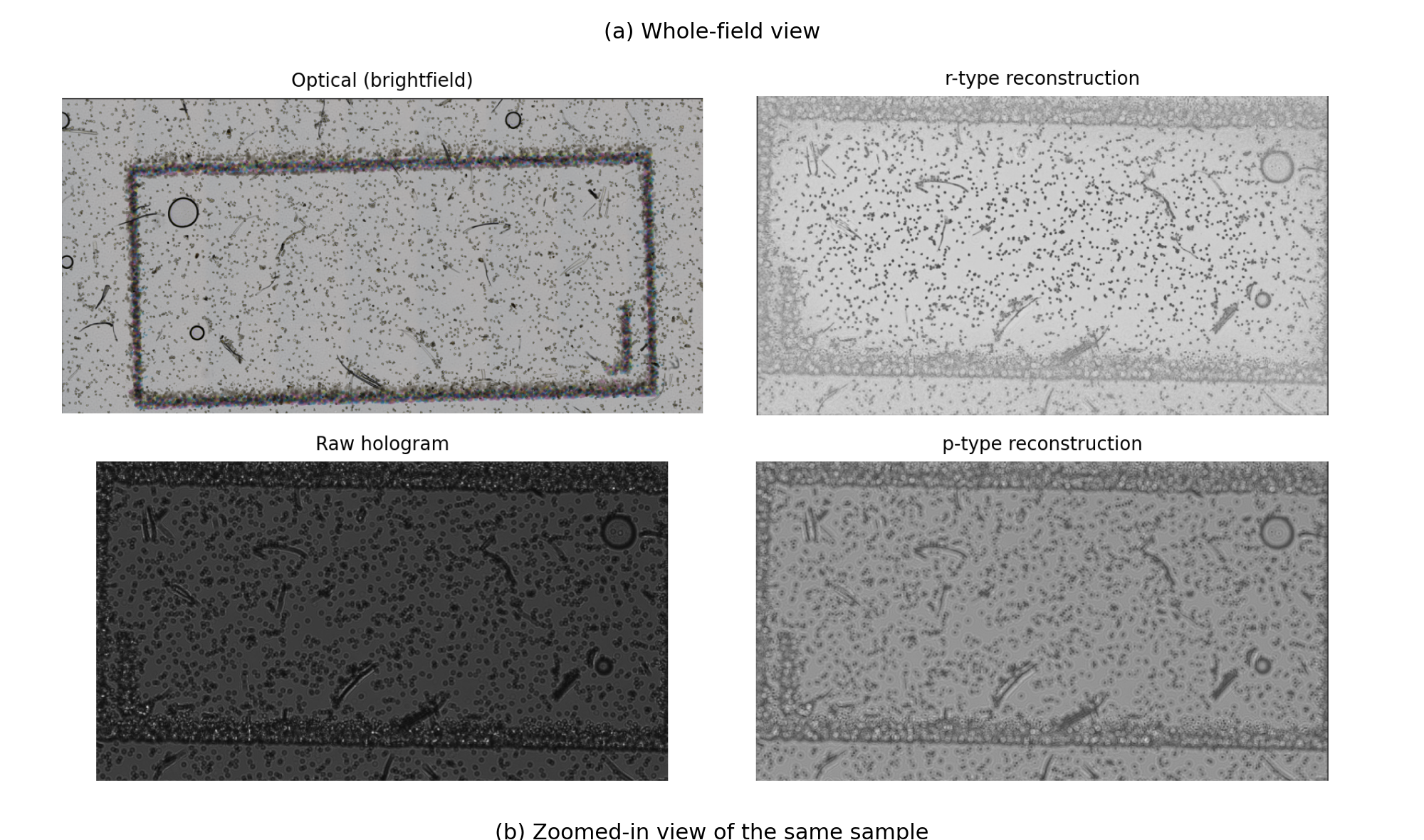}
\caption{\textbf{Whole-field views of the four effective imaging modalities for a single pollen sample.} Optical/brightfield, raw hologram, p-type and r-type reconstructions of the same field of view; the zoomed views are Fig~3 of the main text.}
\label{fig:S_modalities_whole}
\end{figure}

\begin{figure}[H]
\centering
\includegraphics[width=\linewidth]{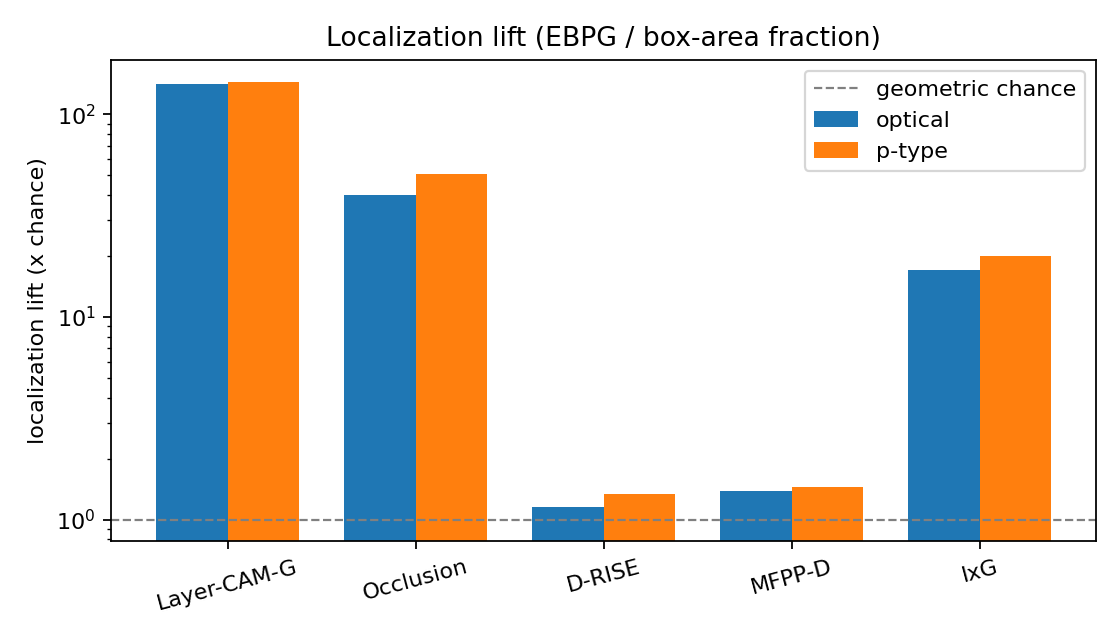}
\caption{\textbf{Localization lift by method and modality.}
In-box energy as a multiple of geometric chance (log scale); the dashed line marks the
random-map level.}
\label{fig:xai_lift}
\end{figure}

\begin{table}[H]
\centering
\caption{\textbf{Decomposition of the Layer-CAM-G localization score.} In-box energy (energy-based pointing game) of plain Layer-CAM without the Gaussian term and of two geometric control maps, against manual ground truth, with 95\% bootstrap intervals; chance is the ground-truth box's share of the image.}
\begin{adjustbox}{max width=\linewidth}\begin{tabular}{|C{0.8in}|C{0.4in}|C{1.4in}|C{1.4in}|C{1.4in}|C{0.55in}|}
\hline
\textbf{Modality} & \textbf{n} & \textbf{Plain Layer-CAM} & \textbf{Gaussian only} & \textbf{Box only} & \textbf{Chance}\\ \hline
Optical & 372 & 0.966 [0.954, 0.976] & 0.730 [0.722, 0.736] & 0.897 [0.886, 0.907] & 0.0067\\ \hline
P-type & 306 & 0.991 [0.986, 0.995] & 0.705 [0.698, 0.712] & 0.838 [0.826, 0.849] & 0.0066\\ \hline
\end{tabular}\end{adjustbox}
\label{table:decomp}
\end{table}

\begin{table}[H]
\centering
\caption{\textbf{D-Deletion and D-Insertion at $\gamma$~=~0.45.} Means over 376 optical and 303 p-type matched detections; D-Deletion lower is better, D-Insertion higher is better. Smoothed maps use $\sigma$~=~16.}
\begin{adjustbox}{max width=\linewidth}\begin{tabular}{|C{1.15in}|C{0.9in}|C{0.9in}|C{0.9in}|C{0.9in}|}
\hline
\textbf{Map} & \textbf{D-Del. optical} & \textbf{D-Ins. optical} & \textbf{D-Del. p-type} & \textbf{D-Ins. p-type}\\ \hline
Random & 0.117 & 0.140 & 0.011 & 0.011\\ \hline
IxG & 0.019 & 0.605 & 0.010 & 0.077\\ \hline
IxG, smoothed & 0.013 & 0.712 & 0.009 & 0.518\\ \hline
Layer-CAM & 0.019 & 0.694 & 0.044 & 0.507\\ \hline
Layer-CAM, smoothed & 0.013 & 0.746 & 0.009 & 0.607\\ \hline
\end{tabular}\end{adjustbox}
\label{table:dfamily}
\end{table}

\begin{table}[H]
\centering
\caption{\textbf{Noise-gate dose response.} Median score retention and fraction of items with zero retention under Gaussian pixel noise, five draws per item, for the detection arms; the operating point of the main text is $\sigma$~=~8/255.}
\begin{tabular}{|l|c|c|c|c|c|c|}
\hline
\textbf{Arm} & \textbf{n} & \textbf{2/255} & \textbf{4/255} & \textbf{8/255} & \textbf{16/255} & \textbf{32/255}\\ \hline
Detection, optical & 372 & 0.992 (0\%) & 0.954 (4\%) & 0.706 (18\%) & 0.000 (64\%) & 0.000 (100\%)\\ \hline
Detection, p-type & 306 & 0.753 (16\%) & 0.000 (100\%) & 0.000 (100\%) & 0.000 (100\%) & 0.000 (100\%)\\ \hline
\end{tabular}
\label{table:S_dose}
\end{table}
Cells give median retention with the percentage of fully collapsed items in parentheses.

\begin{table}[H]
\centering
\caption{\textbf{D-Deletion and D-Insertion at $\gamma$~=~0.50.} Means over the same 376 optical and 303 p-type matched detections as the $\gamma$~=~0.45 values of the main text; the ordering of maps is unchanged.}
\begin{tabular}{|l|c|c|c|c|}
\hline
\textbf{Map} & \textbf{D-Del. optical} & \textbf{D-Ins. optical} & \textbf{D-Del. p-type} & \textbf{D-Ins. p-type}\\ \hline
Random & 0.116 & 0.140 & 0.011 & 0.011\\ \hline
IxG & 0.019 & 0.604 & 0.010 & 0.077\\ \hline
IxG, smoothed & 0.013 & 0.708 & 0.009 & 0.518\\ \hline
Layer-CAM & 0.019 & 0.668 & 0.041 & 0.477\\ \hline
Layer-CAM, smoothed & 0.013 & 0.746 & 0.009 & 0.607\\ \hline
\end{tabular}
\label{table:S_gamma50}
\end{table}

\begin{table}[H]
\centering
\caption{\textbf{Replication of D-Insertion on a seeded subsample.} Mean D-Insertion ($\gamma$~=~0.45) from the full-scale run (376 optical and 303 p-type matched detections) and from an independent run on a seeded random subsample of 50 detections per modality.}
\begin{adjustbox}{max width=\linewidth}
\begin{tabular}{|l|c|c|c|c|}
\hline
\textbf{Map} & \textbf{Optical, full} & \textbf{Optical, subsample} & \textbf{P-type, full} & \textbf{P-type, subsample}\\ \hline
Random & 0.140 & 0.152 & 0.011 & 0.011\\ \hline
IxG & 0.605 & 0.643 & 0.077 & 0.079\\ \hline
IxG, smoothed & 0.712 & 0.748 & 0.518 & 0.548\\ \hline
Layer-CAM & 0.694 & 0.728 & 0.507 & 0.531\\ \hline
Layer-CAM, smoothed & 0.746 & 0.799 & 0.607 & 0.634\\ \hline
\end{tabular}
\end{adjustbox}
\label{table:S_replication}
\end{table}

\begin{table}[H]
\centering
\caption{\textbf{Gate validation across evaluated arms.} Over-all score of the plain IxG map for items that pass the noise gate (retention $\geq$~0.5) and items that fail it, with 95\% bootstrap intervals.}
\begin{tabular}{|l|c|c|c|c|}
\hline
\textbf{Arm} & \textbf{n pass} & \textbf{OA, pass} & \textbf{n fail} & \textbf{OA, fail}\\ \hline
Detection, optical & 228 & +0.674 [+0.650, +0.697] & 144 & +0.442 [+0.408, +0.474]\\ \hline
Detection, p-type & 0 & no items & 306 & +0.054 [+0.049, +0.060]\\ \hline
Classification, optical & 87 & +0.050 [+0.035, +0.065] & 203 & -0.001 [-0.009, +0.007]\\ \hline
Classification, r-type & 148 & -0.024 [-0.033, -0.014] & 135 & -0.006 [-0.012, -0.001]\\ \hline
\end{tabular}
\label{table:S_gatevalid}
\end{table}

\FloatBarrier
\FloatBarrier
\section*{S6. Species-conditioned brittleness}
Brittleness is a property of species features rather than of confidence. On optical classification crops, at an identical clean probability of 0.916, A1 crops retain their full score under noise (retention 1.00) while A2 to A5 crops collapse to about 0.02; on optical detection, retention tracks clean confidence closely (Spearman $\rho$~=~0.75, $p$~=~2.9$\times$10$^{-68}$) with A1 as the exception, whose detections combine a high clean confidence (median 0.898) with low retention (0.110). A gate that operates per item is therefore necessary, and, as the validation above shows, sufficient.

\end{document}